\documentclass{article}
\usepackage{iclr2027_conference,times}

\usepackage{graphicx} \graphicspath{{figure/}}
\usepackage{subcaption}
\usepackage{caption}
\usepackage{overpic,wrapfig}
\usepackage{setspace}

\usepackage{amsmath,amssymb,mathtools}
\usepackage{siunitx}

\usepackage[linesnumbered,ruled,vlined]{algorithm2e}

\usepackage{acronym}
\usepackage{enumitem}
\usepackage{balance}
\usepackage{xspace}

\usepackage[dvipsnames,svgnames,x11names,table]{xcolor}
\usepackage{soul}
\usepackage{booktabs,tabularx,multirow,multicol,makecell}
\usepackage[normalem]{ulem}

\usepackage{pifont}

\usepackage{hyperref}
\usepackage[capitalise,noabbrev,nameinlink]{cleveref}

\makeatletter
\DeclareRobustCommand{\onedot}{\futurelet\@let@token\@onedot}
\newcommand{\@onedot}{\ifx\@let@token.\else.\null\fi\xspace}
\makeatother

\crefname{algorithm}{Alg.}{Algs.}
\Crefname{algocf}{Algorithm}{Algorithms}
\crefname{section}{Sec.}{Secs.}
\Crefname{section}{Section}{Sections}
\crefname{table}{Tab.}{Tabs.}
\Crefname{table}{Table}{Tables}
\crefname{figure}{Fig.}{Figs.}
\Crefname{figure}{Figure}{Figures}
\crefname{equation}{Eq.}{Eqs.}
\Crefname{equation}{Equation}{Equations}
\crefname{appendix}{Appx.}{Appxs.}
\Crefname{appendix}{Appendix}{Appendices}

\DeclareMathOperator*{\argmin}{arg\,min}
\DeclareMathOperator*{\softmax}{softmax}

\definecolor{oursgray}{gray}{0.94}

\acrodef{method}[SilHand]{Hand in Silhouette}
\acrodef{dof}[DoF]{degree of freedom}
\acrodefplural{dof}[DoFs]{degrees of freedom}
\acrodef{mpvpe}[MPVPE]{mean per-vertex position error}
\acrodef{mpjpe}[MPJPE]{mean per-joint position error}
\acrodef{iou}[IoU]{Intersection over Union}

\title{Hand in Silhouette: Synthetic-to-Real Multi-View Hand Reconstruction via Binary Masks}

\author{Leiyao Cui, Lecheng Ruan\textsuperscript{*}, Zihang Zhao\textsuperscript{*}\\
Future Lab, HEC Group\\
\url{https://silhand26.github.io}
}

\iclrfinalcopy 

\begin{document}

\maketitle

\begingroup
\renewcommand{\thefootnote}{\fnsymbol{footnote}}
\footnotetext[1]{Corresponding authors.}
\endgroup

\begin{abstract}
Calibrated multi-view hand reconstruction has advanced rapidly, yet the effectiveness of existing methods depends strongly on real captured data, which is costly to acquire, limited in hand-geometry diversity, and often affected by imperfect annotations. Synthetic data offers a scalable alternative with accurate supervision, but a substantial synthetic-to-real appearance gap hinders its use. To address this, we introduce \textbf{\ac{method}}, a calibrated multi-view hand reconstruction framework that operates on binary hand masks---obtainable at low cost either automatically or manually---to reduce the synthetic-to-real observation gap. Rather than relying on RGB features for 3D lifting, \ac{method} projects metric 3D hypotheses into each view to query and fuse mask-ray evidence for hand reconstruction. This mask-centric formulation enables training on a large-scale synthetic dataset comprising 46.8k grasping poses and 60k free-hand samples, providing broad coverage of hand geometry while directly transferring to real captured hand geometry. Across five real-world benchmarks, \ac{method} improves relative reconstruction accuracy from \textbf{49.2\,\%} to \textbf{63.4\,\%} using masks rendered from reference hands. Moreover, reconstruction accuracy improves consistently as the synthetic training set grows, indicating that broader geometric coverage translates into better real-world performance. While achieving such diversity is difficult with real captured data, it becomes practical under our mask-centric formulation. Finally, \ac{method} remains robust under mask corruption, with reconstruction error increasing by at most \qty{1.63}{\milli\metre}. Together, these results support binary masks as an effective interface for learning hand reconstruction from synthetic geometry.
\end{abstract}

\section{Introduction}

\begin{wrapfigure}{r}{0.25\linewidth}
    \vspace{-1em}

    \centering
    \includegraphics[width=\linewidth]{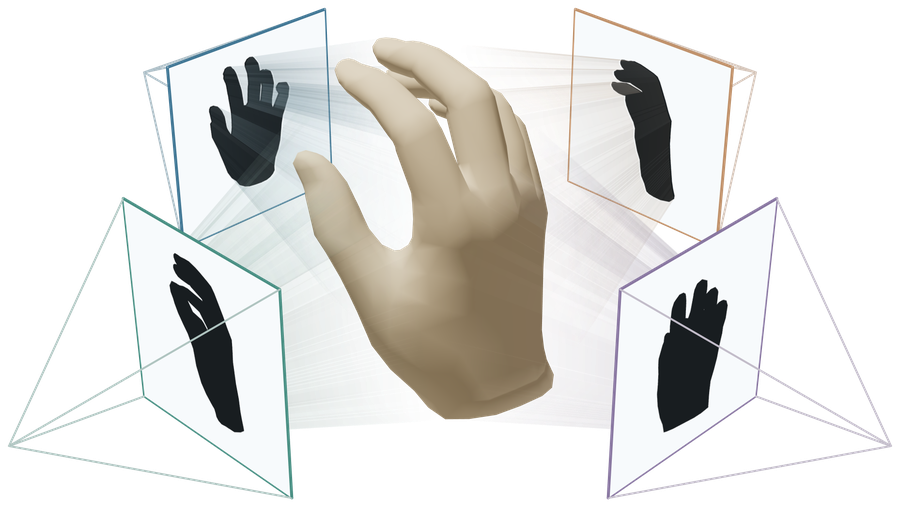}
    \caption{Multi-view masks contain hand geometry cues.}
    \label{fig:teaser}

    \vspace{-0.5em}
\end{wrapfigure}

Accurate 3D hand reconstruction is essential for understanding human–object interaction and sensorimotor control, with broad applications in motion capture~\citep{han2018online,zhang2025hawor}, immersive interaction~\citep{han2020megatrack,wang2025towards}, and embodied intelligence~\citep{zhao2025embedding,cui2026vi,luo2026t}. Calibrated multi-view observations are particularly attractive for this task, as they provide complementary cues that help alleviate ambiguity and occlusion inherent in monocular views~\citep{wehrbein2021probabilistic,zhao2025tac,Wang2026fmpose3d}. Accordingly, recent methods have increasingly exploited cross-view information to improve reconstruction accuracy~\citep{chen2021mvhm,han2022umetrack,yang2023poem,yang2024mlphand,yang2025multi,kuang2026jump}. Nevertheless, achieving robust reconstruction across varied hand geometries remains challenging.

A key bottleneck is training-data diversity and supervision quality. Real multi-camera datasets provide realistic observations~\citep{zimmermann2019freihand,hampali2020honnotate,moon2020interhand,chao2021dexycb,yang2022oakink,fan2023arctic}, but are expensive to acquire, limited in hand-geometry diversity, and hard to annotate accurately due to the complex manifold spanned by the hand configurations. Synthetic data, by contrast, provides exact ground-truth annotations and can offer substantially greater diversity, enabled by advances in human modeling~\citep{loper2023smpl,romero2017embodied,li2022nimble,ferguson2025mhr} and grasp synthesis~\citep{feix2015grasp,chen2025dexonomy}. Yet generating photorealistic appearance remains difficult, leaving a persistent domain gap between synthetic renderings and real observations~\citep{zhao2025analyzing,si2026anyhand}.

\begin{figure}[!t]
    \centering
    \includegraphics[width=\linewidth]{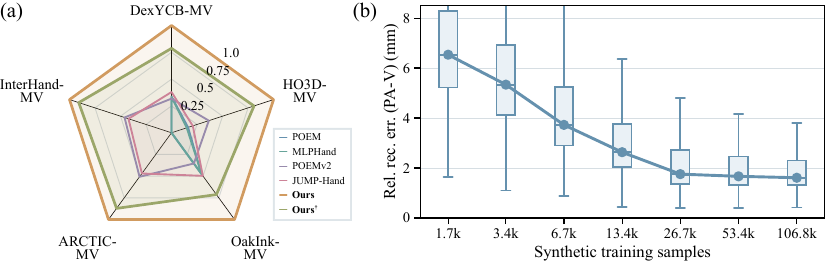}
    \caption{\textbf{Overview of \ac{method}.} (a) Reconstruction result across five real-world multi-view hand benchmarks using reconstruction error (\acs{mpvpe}). Despite being trained on the synthetic dataset, \ac{method} demonstrates strong reconstruction capabilities on real-world geometry. $^{\dagger}$ uses corrupted input masks calibrated to match SAM3 segmentation performance~\citep{carion2026sam}, demonstrating robustness to segmentation noise. (b) Relative reconstruction error (PA-V) as a function of the number of synthetic training samples across the same five datasets.}
    \label{fig:overview}

    \vspace{-15pt}
\end{figure}

In this work, we take a different approach. Rather than making synthetic RGB images increasingly realistic, we remove texture information from the reconstruction input. Specifically, each view is represented by a binary mask (\cref{fig:teaser}), which can be obtained inexpensively and with high accuracy through automatic segmentation or manual annotation~\citep{carion2026sam}. The mask-centric representation discards color, texture, illumination, and background while retaining the projected geometry of the hand~\citep{laurentini1994visual}. Under known cameras, multiple masks provide complementary constraints on the same 3D hand geometry (\cref{fig:teaser}), making them a natural interface between synthetically generated hand geometry and real RGB observations.

Based on this idea, we introduce \ac{method}, a calibrated multi-view hand reconstruction framework operating exclusively on binary hand masks. Without relying on multi-view RGB features for lifting, \ac{method} treats points on a metric 3D grid as hypotheses, projects them into all calibrated views, and queries the corresponding mask-ray features. These view-specific cues are fused with cross-view attention into a shared 3D representation, from which the model estimates global hand pose, canonicalizes the representation, and decodes hand articulation and shape.

This mask-centric formulation enables \ac{method} to be trained on a diverse synthetic dataset of hand geometries, comprising 46.8k grasping poses and 60k free-hand samples. When evaluated directly on five real-world benchmarks---HO3D~\citep{hampali2020honnotate}, InterHand2.6M~\citep{moon2020interhand}, DexYCB~\citep{chao2021dexycb}, OakInk~\citep{yang2022oakink}, and ARCTIC~\citep{fan2023arctic}---\ac{method} improves reconstruction accuracy from \qtyrange{49.2}{63.4}{\%} using masks rendered from reference hands (\cref{fig:overview}(a)). Moreover, performance improves consistently with the scale of synthetic training data (\cref{fig:overview}(b)), indicating that broader geometric training coverage translates into better real-world reconstruction performance. While achieving comparable diversity with real-world hands is challenging, it becomes practical under our mask-centric formulation. The method also remains robust to mask corruption, with reconstruction error increasing by at most \qty{1.63}{\milli\metre} (\cref{fig:overview}(a)).

Our contributions are threefold:
\begin{itemize}
    \vspace{-2pt}
    \item \textbf{A mask-centric synthetic-to-real interface.} We use binary hand masks as an inexpensive, appearance-free interface between synthetic hand geometry and real multi-view observations, avoiding the need for photorealistic synthesis.
    \vspace{-2pt}
    \item \textbf{A multi-view mask-centric reconstruction architecture.} \ac{method} projects metric 3D points into calibrated views to query and fuse mask-ray evidence, followed by global-to-canonical decoding, enabling full 3D hand reconstruction from textureless mask alone.
    \vspace{-2pt}
    \item \textbf{Scalable synthetic geometric supervision.} By varying hand pose, shape, interaction, viewpoint, and mask quality independently, \ac{method} enables large-scale synthetic training that transfers to real datasets and improves as geometric training coverage grows.
\end{itemize}

\section{Related Work}

\paragraph{Multi-View Hand Reconstruction.}
Multi-view hand reconstruction exploits complementary observations to alleviate the depth, pose, and occlusion ambiguities of monocular estimation. Recent learning-based methods have therefore developed increasingly effective strategies for cross-view information aggregation. MVHM~\citep{chen2021mvhm} fuses per-view hand estimates through a learned graph representation, while POEM~\citep{yang2023poem} lifts multi-view features into a 3D representation using keypoints acquired from RGB images. MLPHand~\citep{yang2024mlphand} further introduces an efficient architecture for real-time multi-view estimation. Other directions include collaborative self-supervised learning~\citep{zheng2023hamuco}, training across heterogeneous datasets and camera configurations~\citep{yang2025multi}, and uncertainty-aware fusion under unreliable observations~\citep{kuang2026jump}. Collectively, these approaches improve how 3D information is extracted and integrated from multi-view RGB images. In contrast, we reconsider the observation itself and reconstruct hands from textureless binary masks alone. Rather than relying on RGB features for lifting, we project metric 3D hypotheses into each view to query mask and ray evidence, reducing the synthetic-to-real observation gap and enabling scalable geometric diversity in training.

\paragraph{Real and Synthetic Data for Hand Reconstruction.}
Advances in hand reconstruction have been supported by increasingly large and diverse real-world datasets. FreiHAND~\citep{zimmermann2019freihand} provides supervision for hand pose and shape, while InterHand2.6M~\citep{moon2020interhand} expands coverage of interacting hands. HO3D~\citep{hampali2020honnotate}, DexYCB~\citep{chao2021dexycb}, OakInk~\citep{yang2022oakink}, and ARCTIC~\citep{fan2023arctic} further capture hand-object interactions with diverse objects, viewpoints, and occlusion patterns, with recent datasets continuing to broaden activities and capture settings~\citep{banerjee2025hot3d,fu2025gigahands,rim2026show3d}. Despite this progress, obtaining accurate 3D hand annotations remains costly, making it difficult to densely sample the combinatorial variation in hand shape, articulation, viewpoint, interaction, and visibility.

Synthetic data provides a complementary means of increasing data diversity while offering accurate ground-truth supervision. They have been widely used for hand pose estimation~\citep{zimmermann2017learning,mueller2018ganerated}, hand-object and interacting-hand reconstruction~\citep{hasson2019learning,li2023renderih}, and multi-view estimation~\citep{chen2021mvhm,tse2023spectral}. Recent studies have further investigated how models trained on synthetic data transfer to real hand imagery~\citep{zhao2025analyzing,si2026anyhand}. Nevertheless, a domain gap persists between synthetic and real RGB observations. Our mask-centric formulation sidesteps this issue: synthetic observations need only reproduce projected hand geometry rather than appearance. This enables diverse synthetic hand configurations to transfer to real observations without photorealistic rendering.

\paragraph{Mask-Based 3D Reconstruction.}
Binary masks, traditionally interpreted as object silhouettes, provide a classical representation of multi-view geometric constraints. Shape-from-silhouette and visual-hull methods recover 3D structure by intersecting the viewing cones induced by calibrated foreground masks~\citep{laurentini1994visual,scharr2017fast}, demonstrating that object geometry can be constrained without relying on texture or image correspondences. This principle has also been explored for 3D hand reconstruction. Early model-based approaches estimate hand poses from multiple masks~\citep{ueda2001hand}, while subsequent learning-based methods show that they can support hand reconstruction~\citep{lee2019silhouette,chang2023mask2hand}. Building on this geometric principle, we use binary masks as an appearance-invariant interface between synthetic hand geometry and real multi-view observations, enabling large-scale synthetic training with direct transfer to real captured hand geometry.

\section{\acs{method}}

\subsection{Preliminaries}

Following the parametric hand model convention, MANO hand~\citep{romero2017embodied}, we parameterize the 3D hand geometry by a structured state
\begin{equation}
    \mathcal{H} = (T,\boldsymbol{\theta},\boldsymbol{\beta}),
\end{equation}
where $T\in\mathrm{SE}{(3)}$ denotes the global hand pose, $\boldsymbol{\theta}\in\mathbb{R}^{N_{\boldsymbol{\theta}}}$ the hand pose parameters, and $\boldsymbol{\beta}\in\mathbb{R}^{N_{\boldsymbol{\beta}}}$ the hand shape parameters.

We define the canonical hand space $\{c\}$ as a wrist-centered coordinate frame in which the global rigid transformation is removed. A MANO-based hand model $h$ deterministically maps the intrinsic hand state to the posed geometry in this canonical space
\begin{equation}
    (\mathcal{V}^\text{c}, \mathcal{J}^\text{c}) = h(\boldsymbol{\theta},\boldsymbol{\beta}),
    \label{eq:hand_model}
\end{equation}
where $\mathcal{V}^\text{c}=\{\boldsymbol{v}_i^\text{c}\in\mathbb{R}^3\}_{i=1}^{N_{\mathcal{V}}}$ and $\mathcal{J}^\text{c}=\{\boldsymbol{j}_i^\text{c}\in\mathbb{R}^3\}_{i=1}^{N_{\mathcal{J}}}$ denote the canonical-space mesh vertices and hand joints. The global hand pose $T$ transforms this geometry into the world coordinate frame $\{w\}$:
\begin{equation}
    \mathcal{V}^{\text{w}} = T(\mathcal{V}^{\text{c}}), \quad \mathcal{J}^{\text{w}} = T(\mathcal{J}^{\text{c}}).
    \label{eq:hand_transformation}
\end{equation}
The MANO hand prescribes the triangular face connectivity $\mathcal{F}$, yielding the world-space hand mesh $\mathcal{M} = (\mathcal{V}^\text{w}, \mathcal{F})$. Unless otherwise stated, we omit the world-space superscript $\text{w}$ for brevity.

\begin{figure}[t!]
    \centering
    \includegraphics[width=\linewidth]{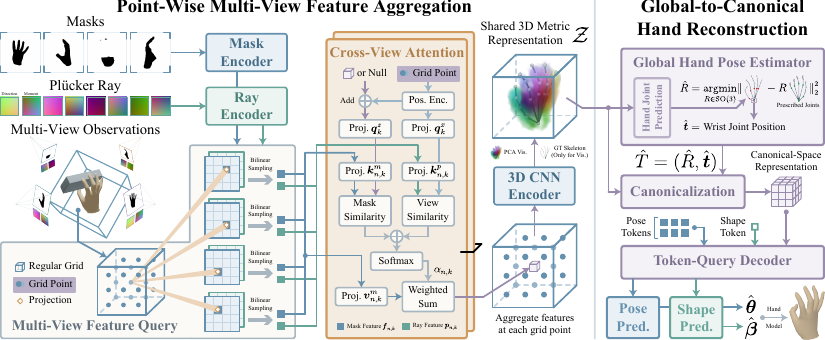}
    \caption{\textbf{Framework of \ac{method}.} Given calibrated multi-view binary masks, \ac{method} projects points from a shared metric 3D grid into each view to query mask and Pl\"ucker-ray features, which are fused by cross-view attention and a 3D CNN into a shared 3D representation. From this representation, \ac{method} first estimates the global hand pose, canonicalizes the 3D evidence, and then uses a token-query decoder to predict the hand pose and shape parameters.}
    \label{fig:framework}
\end{figure}

\subsection{Problem Formulation}

Given $N$ synchronized binary hand masks $\left\{B_n\in\{0, 1\}^{H \times W}\right\}_{n=1}^{N}$ and the corresponding known camera rig $\{\mathcal{C}_n=(K_n, E_n)\}_{n=1}^{N}$, where $K_n\in\mathbb{R}^{3 \times 3}$ and $E_n\in\mathrm{SE}(3)$ denote the camera intrinsics and extrinsics, respectively, we denote the calibrated multi-view input as
\begin{equation}
    \mathcal{O} = \left\{(B_n, \mathcal{C}_n)\right\}_{n=1}^{N}.
\end{equation}

A binary hand mask is determined jointly by the underlying 3D hand state and its view-dependent visibility. Let $\mathcal{Q}$ denote an unknown occlusion configuration. The observed binary mask is determined by a visibility-aware rendering operator $g$:
\begin{equation}
    B_n = g{(\mathcal{H}; \mathcal{Q}, \mathcal{C}_n)},
\end{equation}
Generally, a single mask does not uniquely determine $\mathcal{H}$, as distinct 3D hand configurations can induce identical or similar silhouettes, particularly under occlusion. With calibrated multi-view observations, however, each view imposes an additional geometric constraint on the same underlying hand state. The set of hand states consistent with the multi-view observations is 
\begin{equation}
    \mathbb{H}_{\mathcal{O}} = \bigcap_{n=1}^{N}{\left\{ \mathcal{H}\in\mathbb{H} \mid g{(\mathcal{H}; \mathcal{Q}, \mathcal{C}_n)} = B_n \right\}},
    \label{eq:info_structure}
\end{equation}
where $\mathbb{H}$ denotes the space of valid hand states. While each individual mask admits substantial ambiguity, intersecting the constraints from multiple calibrated views progressively narrows the feasible set, with $\mathbb{H}$ restricting this set to be anatomically and kinematically valid.

We therefore formulate multi-view hand reconstruction as estimating a structured hand state that is consistent with the available geometric evidence:
\begin{equation}
    \hat{\mathcal{H}} = f_{\boldsymbol{\phi}}{(\mathcal{O})}, \quad\hat{\mathcal{H}}\in\mathbb{H}_{\mathcal{O}},
\end{equation}
where $f_{\boldsymbol{\phi}}$ denotes the reconstruction function implemented by \ac{method}. To exploit the geometric structure of the multi-view observations (\cref{eq:info_structure}), we factor the reconstruction into two stages: 
\begin{equation}
    \mathcal{Z} = \mathcal{G}_{\boldsymbol{\phi}}{(\mathcal{O})}, \quad \hat{\mathcal{H}}=\mathcal{R}_{\boldsymbol{\phi}}{(\mathcal{Z})},
    \label{eq:latent}
\end{equation}
where $\mathcal{G}_{\boldsymbol{\phi}}$ uses camera geometry to fuse multi-view mask evidence into a shared metric 3D representation $\mathcal{Z}$, which encodes the features of the feasible set $\mathbb{H}_{\mathcal{O}}$. From this representation, $\mathcal{R}_{\boldsymbol{\phi}}$ infers the structured hand state $\hat{\mathcal{H}}$. The overall framework is illustrated in \cref{fig:framework} and detailed as follows.

\subsection{Point-Wise Multi-View Feature Aggregation}

We first encode each binary mask and its corresponding Pl\"ucker-ray representation using shared mask and ray encoders~\citep{liu2022convnet,woo2023convnext,zhang2024cameras}, producing feature maps $M_n$ and $P_n$, respectively. To remove dependence on the world coordinate choice, the Pl\"ucker rays are computed in a mask-centric coordinate system (\cref{app:sec:metric_grid}). 

We then discretize the mask-centric workspace into a regular metric grid $\{\boldsymbol{x}_k\}_{k=1}^{K}$. Each grid point is projected into every calibrated view to query the corresponding mask and ray features:
\begin{equation}
    \boldsymbol{u}_{n,k} = \pi_n{(\boldsymbol{x}_k;\mathcal{C}_n)}, \quad \boldsymbol{m}_{n,k} = M_n[\boldsymbol{u}_{n,k}], \quad \boldsymbol{p}_{n,k} = P_n[\boldsymbol{u}_{n,k}],
\end{equation}
where $\pi_n(\cdot)$ represents the camera projection, and $[\cdot]$ denotes bilinear sampling. This associates observations across views, without requiring RGB-based correspondences. For clarity, we omit the query, key, and value projections in the following formulation and refer readers to the full computation graph in \cref{fig:framework} for details.

At each grid point, we fuse the queried features using cross-view attention:
\begin{equation}
     \alpha_{n,k} = \softmax{\bigg(\frac{\overbrace{\langle\boldsymbol{q}_k^{z}, \boldsymbol{k}_{n,k}^{m}\rangle}^{\text{mask similarity}} \  + \overbrace{\langle\boldsymbol{q}_k^{x}, \boldsymbol{k}_{n,k}^{p}\rangle}^{\text{view similarity}}}{\sqrt{2d}}\bigg)}, \quad \boldsymbol{z}_k = \sum_{n=1}^N{\alpha_{n,k} \boldsymbol{v}_{n,k}^{m}}.
\end{equation}
The attention weights jointly capture mask consistency and view geometry, while the aggregated feature remains mask-centric. Repeated attention blocks refine the point-wise features, which are finally processed by a 3D CNN to model spatial interactions and produce the shared 3D representation $\mathcal{Z}$ defined in \cref{eq:latent}.

\subsection{Global-to-Canonical Hand Reconstruction}
\paragraph{Global Hand Pose Estimation.}
To decode intrinsic hand pose and shape in a canonical frame, we first recover the global hand transformation from $\mathcal{Z}$. Instead of relying on an unstructured global feature, we use auxiliary world-space joints as geometric anchors: the rigid palm indicates orientation, the wrist determines translation, and joint supervision encourages $\mathcal{Z}$ to preserve hand structure.

We therefore predict auxiliary world-space joints $\tilde{\mathcal{J}}=\left\{\tilde{\boldsymbol j}_i\right\}_{i=1}^{N_J}$ from $\mathcal{Z}$. Each joint is computed as a weighted average over the metric grid:
\begin{equation}
    \tilde{\boldsymbol{j}}_i = \sum_{k=1}^{K}{(w_{i,k}\boldsymbol{x}_k+b_{i,k})}.
\end{equation}

We form $\tilde{\mathcal{J}}_{\text{palm}}$ from the wrist, the thumb carpometacarpal (CMC) joint, and the metacarpophalangeal (MCP) joints of the remaining four fingers (see \cref{app:sec:global_pose} for details). The global rotation is recovered by aligning this predicted palm with the corresponding canonical MANO palm through orthogonal Procrustes~\citep{umeyama1991least,zhao2026tacman}:
\begin{equation}
    \hat{R} = \argmin_{R\in\mathrm{SO}(3)}\left\|\overline{\tilde{\mathcal{J}}_{\text{palm}}} - R(\overline{\mathcal{J}_{\text{palm}}^{\text{c}}})\right\|_2^2,
    \label{eq:palm_alignment}
\end{equation}
where $\overline{(\cdot)}$ represents centroid centered joint coordinates. Since the canonical frame is wrist-centered, the predicted wrist gives the translation $\hat{\boldsymbol{t}}$, yielding $\hat{T}=(\hat{R}, \hat{\boldsymbol{t}})$.

\paragraph{Canonical Hand State Decoding.}
We then express the spatial support of $\mathcal{Z}$ in the estimated canonical frame to decode global-transformation-invariant hand articulation and shape:
\begin{equation}
    \boldsymbol{x}_k^{\text{c}} = \hat{T}^{-1}(\boldsymbol{x}_k), \quad \boldsymbol{z}_k^{\text{c}} = W_z\boldsymbol{z}_k + \gamma_c(\boldsymbol{x}_k^{\text{c}}),
\end{equation}
where $W_z$ projects the aggregated feature and $\gamma_c(\cdot)$ encodes its canonical coordinate. 

A token-query decoder predicts the intrinsic hand articulation $\hat{\boldsymbol{\theta}}$ and neutral canonical geometry $\hat{\mathcal{V}}^0$, following~\citep{kulon2020weakly,wang2023memahand,yu2023overcoming}. We constrain $\hat{\boldsymbol{\theta}}$ to prescribed anatomical ranges~\citep{yang2021cpf} and project $\hat{\mathcal{V}}^0$ onto the fixed MANO shape basis to obtain $\hat{\boldsymbol{\beta}}$. This fixed projection introduces no additional learned shape parameters and is detailed in \cref{app:sec:shape_projection}. The resulting hand state $\hat{\mathcal{H}}=(\hat{T},\hat{\boldsymbol{\theta}},\hat{\boldsymbol{\beta}})$ reconstructs the final geometry via \cref{eq:hand_model,eq:hand_transformation}.

\subsection{Training Objectives}

Synthetic data provides ground-truth global pose $T^{*}=(R^*,\boldsymbol{t}^*)$, hand articulation $\boldsymbol{\theta}^{*}$, world-space joint positions $\mathcal{J}^*$, and neutral canonical hand geometry $V^{0,*}$. We supervise:
\begin{equation}
\begin{aligned}
    \mathcal{L}_{\text{pose}} = \text{smooth-}\ell_1\left(\hat{\boldsymbol{\theta}}, \boldsymbol{\theta}^*\right),& \quad \mathcal{L}_{\text{shape}} = \text{smooth-}\ell_1\left(\tilde{\mathcal{V}^0}, \mathcal{V}^{0,*}\right), \\
    \quad \mathcal{L}_{\text{joint}} = \text{smooth-}\ell_1\left(\tilde{\mathcal{J}}, \mathcal{J}^*\right),& \quad \mathcal{L}_{\text{rot}} = \left\|\hat R-R^\ast\right\|_\text{F}^2.
\end{aligned}
\end{equation}
The total objective is
\begin{equation}
    \mathcal{L} = \lambda_{\text{pose}} \mathcal{L}_{\text{pose}} + \lambda_{\text{shape}} \mathcal{L}_{\text{shape}} + 
    \lambda_{\text{joint}} \mathcal{L}_{\text{joint}} + \lambda_{\text{rot}} \mathcal{L}_{\text{rot}}.
    \label{eq:total_loss}
\end{equation}
Loss weights and other details are reported in \cref{app:sec:training_details}.

\subsection{Scalable Synthetic Geometric Supervision}\label{sec:data_generation}

\begin{wrapfigure}{r}{0.3\linewidth}
    \vspace{-1.5em}

    \centering
    \includegraphics[width=\linewidth]{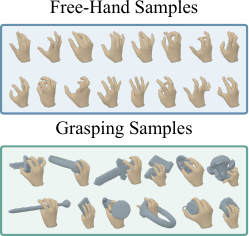}
    \caption{Examples from our synthetic dataset.}
    \label{fig:data_example}

    \vspace{-0.5em}
\end{wrapfigure}

We train with synthetic hand geometry, comprising free-hand and grasping configurations, to capture collision effects. We initially sample free-hand and grasping configurations in a 1:1 ratio. For free-hand configurations, we sample the MANO pose PCA space~\citep{romero2017embodied} and reject samples that violate prescribed anatomical joint ranges or exhibit hand self-collision. For grasping configurations, we adapt Dexonomy~\citep{chen2025dexonomy} to our MANO-based hand-state parameterization. In addition to filtering out hand self-collisions, we perform hand-object collision checking to ensure the geometric validity of the grasps.

In the absence of an established hand-shape distribution~\citep{kulon2020weakly,wang2023memahand,yu2023overcoming}, we sample shape parameters from an empirical distribution fitted to eight datasets~\citep{chao2021dexycb,han2022umetrack,moon2023reinterhand,fan2023arctic,wang2025ho,banerjee2025hot3d,fan2026palm}. After applying the anatomical and collision-based filtering, the final dataset comprises 60k free-hand configurations and 46.8k grasping configurations, with representative examples shown in \cref{fig:data_example}.

Following the calibrated multi-view setting with known camera geometry, we train using randomized camera parameters sampled from the datasets: HO3D~\citep{hampali2020honnotate}, InterHand2.6M~\citep{moon2020interhand}, DexYCB~\citep{chao2021dexycb}, OakInk~\citep{yang2022oakink}, and ARCTIC~\citep{fan2023arctic}, like POEMv2~\citep{yang2025multi}. We perturb the camera parameters to broaden the viewing distribution, then sample global hand poses and render the masks.

To improve robustness to imperfect observations, we perturb the rendered masks by varying hand-region visibility and forearm extent, applying morphological and boundary perturbations, and introducing localized boundary artifacts. We further apply random view dropout for augmentation. All perturbations modify only the observations, while keeping the underlying target unchanged.

\section{Experiments and Results}

\begin{table}[!t]
\centering
\caption{\textbf{Quantitative results of \ac{method}.} $\downarrow$ indicates lower is better. Best results are \textbf{bold}, and the best baseline results are \underline{underlined}. \textcolor{green!60!black}{Green values} indicate the absolute improvement over the best baseline result. \textsubscript{C1}, \textsubscript{C2}, \textsubscript{C3}, and \textsubscript{C4} indicate progressively stronger synthetic mask corruption.}
\label{tab:main_results}

\small

\begin{tabular}{cccccccc}
\toprule
Dataset & Method & MPVPE $\downarrow$ & RR-V $\downarrow$ & PA-V $\downarrow$ & MPJPE $\downarrow$ & RR-J $\downarrow$ & PA-J $\downarrow$ \\ 
\midrule
\multirow{10}{*}{\begin{tabular}[c]{@{}c@{}}DexYCB-\\MV\end{tabular}} & POEM & 6.13 & 7.21 & 4.00 & 6.06 & 7.30 & 3.93 \\
& MLPHand & 6.20 & 7.30 & 4.15 & 6.23 & 7.41 & 4.11 \\
& POEMv2 & 6.50 & 7.50 & 3.84 & 6.33 & 7.54 & 3.68 \\
& JUMP-Hand & \uline{5.45} & \uline{6.41} & \uline{3.77} & \uline{5.31} & \uline{6.41} & \uline{3.61} \\
\cmidrule{2-8}
& Visual Hull & -- & -- & 12.95 & -- & -- & 12.42 \\
& \cellcolor{oursgray}\textbf{Ours}
& \cellcolor{oursgray}\textbf{2.07}\textsubscript{\,\textcolor{green!60!black}{$\downarrow 3.38$}}
& \cellcolor{oursgray}\textbf{1.63}\textsubscript{\,\textcolor{green!60!black}{$\downarrow 4.78$}}
& \cellcolor{oursgray}\textbf{1.12}\textsubscript{\,\textcolor{green!60!black}{$\downarrow 2.65$}}
& \cellcolor{oursgray}\textbf{2.12}\textsubscript{\,\textcolor{green!60!black}{$\downarrow 3.19$}}
& \cellcolor{oursgray}\textbf{1.69}\textsubscript{\,\textcolor{green!60!black}{$\downarrow 4.72$}}
& \cellcolor{oursgray}\textbf{1.16}\textsubscript{\,\textcolor{green!60!black}{$\downarrow 2.45$}} \\
& \cellcolor{oursgray}\textbf{Ours\rlap{\,\textsubscript{C1}}}
& \cellcolor{oursgray}2.18\textsubscript{\,\textcolor{green!60!black}{$\downarrow 3.27$}}
& \cellcolor{oursgray}1.82\textsubscript{\,\textcolor{green!60!black}{$\downarrow 4.59$}}
& \cellcolor{oursgray}1.23\textsubscript{\,\textcolor{green!60!black}{$\downarrow 2.54$}}
& \cellcolor{oursgray}2.23\textsubscript{\,\textcolor{green!60!black}{$\downarrow 3.08$}}
& \cellcolor{oursgray}1.89\textsubscript{\,\textcolor{green!60!black}{$\downarrow 4.52$}}
& \cellcolor{oursgray}1.27\textsubscript{\,\textcolor{green!60!black}{$\downarrow 2.34$}} \\
& \cellcolor{oursgray}\textbf{Ours\rlap{\,\textsubscript{C2}}}
& \cellcolor{oursgray}2.41\textsubscript{\,\textcolor{green!60!black}{$\downarrow 3.04$}}
& \cellcolor{oursgray}2.01\textsubscript{\,\textcolor{green!60!black}{$\downarrow 4.40$}}
& \cellcolor{oursgray}1.37\textsubscript{\,\textcolor{green!60!black}{$\downarrow 2.40$}}
& \cellcolor{oursgray}2.46\textsubscript{\,\textcolor{green!60!black}{$\downarrow 2.85$}}
& \cellcolor{oursgray}2.08\textsubscript{\,\textcolor{green!60!black}{$\downarrow 4.33$}}
& \cellcolor{oursgray}1.42\textsubscript{\,\textcolor{green!60!black}{$\downarrow 2.19$}} \\
& \cellcolor{oursgray}\textbf{Ours\rlap{\,\textsubscript{C3}}}
& \cellcolor{oursgray}2.63\textsubscript{\,\textcolor{green!60!black}{$\downarrow 2.82$}}
& \cellcolor{oursgray}2.15\textsubscript{\,\textcolor{green!60!black}{$\downarrow 4.26$}}
& \cellcolor{oursgray}1.49\textsubscript{\,\textcolor{green!60!black}{$\downarrow 2.28$}}
& \cellcolor{oursgray}2.68\textsubscript{\,\textcolor{green!60!black}{$\downarrow 2.63$}}
& \cellcolor{oursgray}2.23\textsubscript{\,\textcolor{green!60!black}{$\downarrow 4.18$}}
& \cellcolor{oursgray}1.55\textsubscript{\,\textcolor{green!60!black}{$\downarrow 2.06$}} \\
& \cellcolor{oursgray}\textbf{Ours\rlap{\,\textsubscript{C4}}}
& \cellcolor{oursgray}2.98\textsubscript{\,\textcolor{green!60!black}{$\downarrow 2.47$}}
& \cellcolor{oursgray}2.38\textsubscript{\,\textcolor{green!60!black}{$\downarrow 4.03$}}
& \cellcolor{oursgray}1.63\textsubscript{\,\textcolor{green!60!black}{$\downarrow 2.14$}}
& \cellcolor{oursgray}3.04\textsubscript{\,\textcolor{green!60!black}{$\downarrow 2.27$}}
& \cellcolor{oursgray}2.47\textsubscript{\,\textcolor{green!60!black}{$\downarrow 3.94$}}
& \cellcolor{oursgray}1.69\textsubscript{\,\textcolor{green!60!black}{$\downarrow 1.92$}} \\ 
\midrule
\multirow{10}{*}{\begin{tabular}[c]{@{}c@{}}HO3D-\\MV\end{tabular}} & POEM & 17.20 & 21.45 & 9.97 & 17.28 & 21.94 & 9.60 \\
& MLPHand & 18.69 & 23.28 & 10.54 & 18.70 & 23.76 & 10.11 \\
& POEMv2 & \uline{7.63} & \uline{10.58} & \uline{4.59} & \uline{7.18} & \uline{10.58} & \uline{4.13} \\
& JUMP-Hand & 13.39 & 16.96 & 8.78 & 13.10 & 17.07 & 8.30 \\
\cmidrule{2-8}
& Visual Hull & -- & -- & 10.56 & -- & -- & 10.22 \\
& \cellcolor{oursgray}\textbf{Ours}
& \cellcolor{oursgray}\textbf{2.79}\textsubscript{\,\textcolor{green!60!black}{$\downarrow 4.84$}}
& \cellcolor{oursgray}\textbf{2.70}\textsubscript{\,\textcolor{green!60!black}{$\downarrow 7.88$}}
& \cellcolor{oursgray}\textbf{1.83}\textsubscript{\,\textcolor{green!60!black}{$\downarrow 2.76$}}
& \cellcolor{oursgray}\textbf{2.76}\textsubscript{\,\textcolor{green!60!black}{$\downarrow 4.42$}}
& \cellcolor{oursgray}\textbf{2.67}\textsubscript{\,\textcolor{green!60!black}{$\downarrow 7.91$}}
& \cellcolor{oursgray}\textbf{1.76}\textsubscript{\,\textcolor{green!60!black}{$\downarrow 2.37$}} \\
& \cellcolor{oursgray}\textbf{Ours\rlap{\,\textsubscript{C1}}}
& \cellcolor{oursgray}3.03\textsubscript{\,\textcolor{green!60!black}{$\downarrow 4.60$}}
& \cellcolor{oursgray}2.95\textsubscript{\,\textcolor{green!60!black}{$\downarrow 7.63$}}
& \cellcolor{oursgray}1.89\textsubscript{\,\textcolor{green!60!black}{$\downarrow 2.70$}}
& \cellcolor{oursgray}3.01\textsubscript{\,\textcolor{green!60!black}{$\downarrow 4.17$}}
& \cellcolor{oursgray}2.92\textsubscript{\,\textcolor{green!60!black}{$\downarrow 7.66$}}
& \cellcolor{oursgray}1.83\textsubscript{\,\textcolor{green!60!black}{$\downarrow 2.30$}} \\
& \cellcolor{oursgray}\textbf{Ours\rlap{\,\textsubscript{C2}}}
& \cellcolor{oursgray}3.31\textsubscript{\,\textcolor{green!60!black}{$\downarrow 4.32$}}
& \cellcolor{oursgray}3.16\textsubscript{\,\textcolor{green!60!black}{$\downarrow 7.42$}}
& \cellcolor{oursgray}1.99\textsubscript{\,\textcolor{green!60!black}{$\downarrow 2.60$}}
& \cellcolor{oursgray}3.28\textsubscript{\,\textcolor{green!60!black}{$\downarrow 3.90$}}
& \cellcolor{oursgray}3.15\textsubscript{\,\textcolor{green!60!black}{$\downarrow 7.43$}}
& \cellcolor{oursgray}1.93\textsubscript{\,\textcolor{green!60!black}{$\downarrow 2.20$}} \\
& \cellcolor{oursgray}\textbf{Ours\rlap{\,\textsubscript{C3}}}
& \cellcolor{oursgray}3.45\textsubscript{\,\textcolor{green!60!black}{$\downarrow 4.18$}}
& \cellcolor{oursgray}3.31\textsubscript{\,\textcolor{green!60!black}{$\downarrow 7.27$}}
& \cellcolor{oursgray}2.07\textsubscript{\,\textcolor{green!60!black}{$\downarrow 2.52$}}
& \cellcolor{oursgray}3.42\textsubscript{\,\textcolor{green!60!black}{$\downarrow 3.76$}}
& \cellcolor{oursgray}3.30\textsubscript{\,\textcolor{green!60!black}{$\downarrow 7.28$}}
& \cellcolor{oursgray}2.01\textsubscript{\,\textcolor{green!60!black}{$\downarrow 2.12$}} \\
& \cellcolor{oursgray}\textbf{Ours\rlap{\,\textsubscript{C4}}}
& \cellcolor{oursgray}3.55\textsubscript{\,\textcolor{green!60!black}{$\downarrow 4.08$}}
& \cellcolor{oursgray}3.47\textsubscript{\,\textcolor{green!60!black}{$\downarrow 7.11$}}
& \cellcolor{oursgray}2.16\textsubscript{\,\textcolor{green!60!black}{$\downarrow 2.43$}}
& \cellcolor{oursgray}3.52\textsubscript{\,\textcolor{green!60!black}{$\downarrow 3.66$}}
& \cellcolor{oursgray}3.47\textsubscript{\,\textcolor{green!60!black}{$\downarrow 7.11$}}
& \cellcolor{oursgray}2.10\textsubscript{\,\textcolor{green!60!black}{$\downarrow 2.03$}} \\ 
\midrule
\multirow{10}{*}{\begin{tabular}[c]{@{}c@{}}OakInk-\\MV\end{tabular}} & POEM & 6.20 & 7.63 & 4.21 & 6.01 & 7.46 & 4.00 \\
& MLPHand & 6.17 & 7.49 & 4.26 & 6.00 & 7.52 & 4.09 \\
& POEMv2 & 8.33 & 8.47 & 4.49 & 8.09 & 8.47 & 4.24 \\
& JUMP-Hand & \uline{5.94} & \uline{6.99} & \uline{4.13} & \uline{5.72} & \uline{6.94} & \uline{3.86} \\
\cmidrule{2-8}
& Visual Hull & -- & -- & 13.06 & -- & -- & 12.78 \\
& \cellcolor{oursgray}\textbf{Ours}
& \cellcolor{oursgray}\textbf{2.95}\textsubscript{\,\textcolor{green!60!black}{$\downarrow 2.99$}}
& \cellcolor{oursgray}\textbf{2.90}\textsubscript{\,\textcolor{green!60!black}{$\downarrow 4.09$}}
& \cellcolor{oursgray}\textbf{2.25}\textsubscript{\,\textcolor{green!60!black}{$\downarrow 1.88$}}
& \cellcolor{oursgray}\textbf{3.00}\textsubscript{\,\textcolor{green!60!black}{$\downarrow 2.72$}}
& \cellcolor{oursgray}\textbf{2.95}\textsubscript{\,\textcolor{green!60!black}{$\downarrow 3.99$}}
& \cellcolor{oursgray}\textbf{2.25}\textsubscript{\,\textcolor{green!60!black}{$\downarrow 1.61$}} \\
& \cellcolor{oursgray}\textbf{Ours\rlap{\,\textsubscript{C1}}}
& \cellcolor{oursgray}3.31\textsubscript{\,\textcolor{green!60!black}{$\downarrow 2.63$}}
& \cellcolor{oursgray}3.25\textsubscript{\,\textcolor{green!60!black}{$\downarrow 3.74$}}
& \cellcolor{oursgray}2.48\textsubscript{\,\textcolor{green!60!black}{$\downarrow 1.65$}}
& \cellcolor{oursgray}3.37\textsubscript{\,\textcolor{green!60!black}{$\downarrow 2.35$}}
& \cellcolor{oursgray}3.31\textsubscript{\,\textcolor{green!60!black}{$\downarrow 3.63$}}
& \cellcolor{oursgray}2.49\textsubscript{\,\textcolor{green!60!black}{$\downarrow 1.37$}} \\
& \cellcolor{oursgray}\textbf{Ours\rlap{\,\textsubscript{C2}}}
& \cellcolor{oursgray}3.77\textsubscript{\,\textcolor{green!60!black}{$\downarrow 2.17$}}
& \cellcolor{oursgray}3.66\textsubscript{\,\textcolor{green!60!black}{$\downarrow 3.33$}}
& \cellcolor{oursgray}2.76\textsubscript{\,\textcolor{green!60!black}{$\downarrow 1.37$}}
& \cellcolor{oursgray}3.83\textsubscript{\,\textcolor{green!60!black}{$\downarrow 1.89$}}
& \cellcolor{oursgray}3.74\textsubscript{\,\textcolor{green!60!black}{$\downarrow 3.20$}}
& \cellcolor{oursgray}2.78\textsubscript{\,\textcolor{green!60!black}{$\downarrow 1.08$}} \\
& \cellcolor{oursgray}\textbf{Ours\rlap{\,\textsubscript{C3}}}
& \cellcolor{oursgray}4.14\textsubscript{\,\textcolor{green!60!black}{$\downarrow 1.80$}}
& \cellcolor{oursgray}4.01\textsubscript{\,\textcolor{green!60!black}{$\downarrow 2.98$}}
& \cellcolor{oursgray}2.97\textsubscript{\,\textcolor{green!60!black}{$\downarrow 1.16$}}
& \cellcolor{oursgray}4.21\textsubscript{\,\textcolor{green!60!black}{$\downarrow 1.51$}}
& \cellcolor{oursgray}4.10\textsubscript{\,\textcolor{green!60!black}{$\downarrow 2.84$}}
& \cellcolor{oursgray}3.01\textsubscript{\,\textcolor{green!60!black}{$\downarrow 0.85$}} \\
& \cellcolor{oursgray}\textbf{Ours\rlap{\,\textsubscript{C4}}}
& \cellcolor{oursgray}4.58\textsubscript{\,\textcolor{green!60!black}{$\downarrow 1.36$}}
& \cellcolor{oursgray}4.40\textsubscript{\,\textcolor{green!60!black}{$\downarrow 2.59$}}
& \cellcolor{oursgray}3.21\textsubscript{\,\textcolor{green!60!black}{$\downarrow 0.92$}}
& \cellcolor{oursgray}4.66\textsubscript{\,\textcolor{green!60!black}{$\downarrow 1.06$}}
& \cellcolor{oursgray}4.52\textsubscript{\,\textcolor{green!60!black}{$\downarrow 2.42$}}
& \cellcolor{oursgray}3.26\textsubscript{\,\textcolor{green!60!black}{$\downarrow 0.60$}} \\ 
\midrule
\multirow{8}{*}{\begin{tabular}[c]{@{}c@{}}ARCTIC-\\MV\end{tabular}} & POEMv2 & \uline{5.89} & \uline{7.13} & \uline{4.43} & \uline{5.42} & \uline{6.80} & \uline{3.91} \\
& JUMP-Hand & 6.37 & -- & -- & -- & -- & -- \\
\cmidrule{2-8}
& Visual Hull & -- & -- & 15.78 & -- & -- & 15.15 \\
& \cellcolor{oursgray}\textbf{Ours}
& \cellcolor{oursgray}\textbf{2.99}\textsubscript{\,\textcolor{green!60!black}{$\downarrow 2.90$}}
& \cellcolor{oursgray}\textbf{2.97}\textsubscript{\,\textcolor{green!60!black}{$\downarrow 4.16$}}
& \cellcolor{oursgray}\textbf{2.20}\textsubscript{\,\textcolor{green!60!black}{$\downarrow 2.23$}}
& \cellcolor{oursgray}\textbf{3.00}\textsubscript{\,\textcolor{green!60!black}{$\downarrow 2.42$}}
& \cellcolor{oursgray}\textbf{2.99}\textsubscript{\,\textcolor{green!60!black}{$\downarrow 3.81$}}
& \cellcolor{oursgray}\textbf{2.18}\textsubscript{\,\textcolor{green!60!black}{$\downarrow 1.73$}} \\
& \cellcolor{oursgray}\textbf{Ours\rlap{\,\textsubscript{C1}}}
& \cellcolor{oursgray}3.07\textsubscript{\,\textcolor{green!60!black}{$\downarrow 2.82$}}
& \cellcolor{oursgray}3.04\textsubscript{\,\textcolor{green!60!black}{$\downarrow 4.09$}}
& \cellcolor{oursgray}2.26\textsubscript{\,\textcolor{green!60!black}{$\downarrow 2.17$}}
& \cellcolor{oursgray}3.09\textsubscript{\,\textcolor{green!60!black}{$\downarrow 2.33$}}
& \cellcolor{oursgray}3.06\textsubscript{\,\textcolor{green!60!black}{$\downarrow 3.74$}}
& \cellcolor{oursgray}2.25\textsubscript{\,\textcolor{green!60!black}{$\downarrow 1.66$}} \\
& \cellcolor{oursgray}\textbf{Ours\rlap{\,\textsubscript{C2}}}
& \cellcolor{oursgray}3.22\textsubscript{\,\textcolor{green!60!black}{$\downarrow 2.67$}}
& \cellcolor{oursgray}3.15\textsubscript{\,\textcolor{green!60!black}{$\downarrow 3.98$}}
& \cellcolor{oursgray}2.33\textsubscript{\,\textcolor{green!60!black}{$\downarrow 2.10$}}
& \cellcolor{oursgray}3.24\textsubscript{\,\textcolor{green!60!black}{$\downarrow 2.18$}}
& \cellcolor{oursgray}3.18\textsubscript{\,\textcolor{green!60!black}{$\downarrow 3.62$}}
& \cellcolor{oursgray}2.32\textsubscript{\,\textcolor{green!60!black}{$\downarrow 1.59$}} \\
& \cellcolor{oursgray}\textbf{Ours\rlap{\,\textsubscript{C3}}}
& \cellcolor{oursgray}3.43\textsubscript{\,\textcolor{green!60!black}{$\downarrow 2.46$}}
& \cellcolor{oursgray}3.31\textsubscript{\,\textcolor{green!60!black}{$\downarrow 3.82$}}
& \cellcolor{oursgray}2.41\textsubscript{\,\textcolor{green!60!black}{$\downarrow 2.02$}}
& \cellcolor{oursgray}3.45\textsubscript{\,\textcolor{green!60!black}{$\downarrow 1.97$}}
& \cellcolor{oursgray}3.34\textsubscript{\,\textcolor{green!60!black}{$\downarrow 3.46$}}
& \cellcolor{oursgray}2.41\textsubscript{\,\textcolor{green!60!black}{$\downarrow 1.50$}} \\
& \cellcolor{oursgray}\textbf{Ours\rlap{\,\textsubscript{C4}}}
& \cellcolor{oursgray}3.76\textsubscript{\,\textcolor{green!60!black}{$\downarrow 2.13$}}
& \cellcolor{oursgray}3.56\textsubscript{\,\textcolor{green!60!black}{$\downarrow 3.57$}}
& \cellcolor{oursgray}2.55\textsubscript{\,\textcolor{green!60!black}{$\downarrow 1.88$}}
& \cellcolor{oursgray}3.77\textsubscript{\,\textcolor{green!60!black}{$\downarrow 1.65$}}
& \cellcolor{oursgray}3.61\textsubscript{\,\textcolor{green!60!black}{$\downarrow 3.19$}}
& \cellcolor{oursgray}2.55\textsubscript{\,\textcolor{green!60!black}{$\downarrow 1.36$}} \\ 
\midrule
\multirow{8}{*}{\begin{tabular}[c]{@{}c@{}}InterHand-\\MV\end{tabular}} & POEMv2 & \uline{7.64} & \uline{9.04} & \uline{5.23} & \uline{7.19} & \uline{8.82} & \uline{4.72} \\
& JUMP-Hand & 8.34 & -- & -- & -- & -- & -- \\
\cmidrule{2-8}
& Visual Hull & -- & -- & 20.51 & -- & -- & 19.89 \\
& \cellcolor{oursgray}\textbf{Ours}
& \cellcolor{oursgray}\textbf{3.52}\textsubscript{\,\textcolor{green!60!black}{$\downarrow 4.12$}}
& \cellcolor{oursgray}\textbf{3.39}\textsubscript{\,\textcolor{green!60!black}{$\downarrow 5.65$}}
& \cellcolor{oursgray}\textbf{2.18}\textsubscript{\,\textcolor{green!60!black}{$\downarrow 3.05$}}
& \cellcolor{oursgray}\textbf{3.59}\textsubscript{\,\textcolor{green!60!black}{$\downarrow 3.60$}}
& \cellcolor{oursgray}\textbf{3.45}\textsubscript{\,\textcolor{green!60!black}{$\downarrow 5.37$}}
& \cellcolor{oursgray}\textbf{2.17}\textsubscript{\,\textcolor{green!60!black}{$\downarrow 2.55$}} \\
& \cellcolor{oursgray}\textbf{Ours\rlap{\,\textsubscript{C1}}}
& \cellcolor{oursgray}3.67\textsubscript{\,\textcolor{green!60!black}{$\downarrow 3.97$}}
& \cellcolor{oursgray}3.51\textsubscript{\,\textcolor{green!60!black}{$\downarrow 5.53$}}
& \cellcolor{oursgray}2.24\textsubscript{\,\textcolor{green!60!black}{$\downarrow 2.99$}}
& \cellcolor{oursgray}3.75\textsubscript{\,\textcolor{green!60!black}{$\downarrow 3.44$}}
& \cellcolor{oursgray}3.58\textsubscript{\,\textcolor{green!60!black}{$\downarrow 5.24$}}
& \cellcolor{oursgray}2.23\textsubscript{\,\textcolor{green!60!black}{$\downarrow 2.49$}} \\
& \cellcolor{oursgray}\textbf{Ours\rlap{\,\textsubscript{C2}}}
& \cellcolor{oursgray}3.79\textsubscript{\,\textcolor{green!60!black}{$\downarrow 3.85$}}
& \cellcolor{oursgray}3.51\textsubscript{\,\textcolor{green!60!black}{$\downarrow 5.53$}}
& \cellcolor{oursgray}2.29\textsubscript{\,\textcolor{green!60!black}{$\downarrow 2.94$}}
& \cellcolor{oursgray}3.86\textsubscript{\,\textcolor{green!60!black}{$\downarrow 3.33$}}
& \cellcolor{oursgray}3.58\textsubscript{\,\textcolor{green!60!black}{$\downarrow 5.24$}}
& \cellcolor{oursgray}2.29\textsubscript{\,\textcolor{green!60!black}{$\downarrow 2.43$}} \\
& \cellcolor{oursgray}\textbf{Ours\rlap{\,\textsubscript{C3}}}
& \cellcolor{oursgray}3.87\textsubscript{\,\textcolor{green!60!black}{$\downarrow 3.77$}}
& \cellcolor{oursgray}3.53\textsubscript{\,\textcolor{green!60!black}{$\downarrow 5.51$}}
& \cellcolor{oursgray}2.35\textsubscript{\,\textcolor{green!60!black}{$\downarrow 2.88$}}
& \cellcolor{oursgray}3.94\textsubscript{\,\textcolor{green!60!black}{$\downarrow 3.25$}}
& \cellcolor{oursgray}3.60\textsubscript{\,\textcolor{green!60!black}{$\downarrow 5.22$}}
& \cellcolor{oursgray}2.36\textsubscript{\,\textcolor{green!60!black}{$\downarrow 2.36$}} \\
& \cellcolor{oursgray}\textbf{Ours\rlap{\,\textsubscript{C4}}}
& \cellcolor{oursgray}4.00\textsubscript{\,\textcolor{green!60!black}{$\downarrow 3.64$}}
& \cellcolor{oursgray}3.59\textsubscript{\,\textcolor{green!60!black}{$\downarrow 5.45$}}
& \cellcolor{oursgray}2.44\textsubscript{\,\textcolor{green!60!black}{$\downarrow 2.79$}}
& \cellcolor{oursgray}4.06\textsubscript{\,\textcolor{green!60!black}{$\downarrow 3.13$}}
& \cellcolor{oursgray}3.66\textsubscript{\,\textcolor{green!60!black}{$\downarrow 5.16$}}
& \cellcolor{oursgray}2.45\textsubscript{\,\textcolor{green!60!black}{$\downarrow 2.27$}} \\
\bottomrule
\end{tabular}

\end{table}

We evaluate \ac{method} on five real-world datasets: HO3D-MV~\citep{hampali2020honnotate}, InterHand-MV~\citep{moon2020interhand}, DexYCB-MV~\citep{chao2021dexycb}, OakInk-MV~\citep{yang2022oakink}, and ARCTIC-MV~\citep{fan2023arctic}. We compare against four learning-based multi-view reconstruction methods, including POEM~\citep{yang2023poem}, MLPHand~\citep{yang2024mlphand}, POEMv2~\citep{yang2025multi}, and JUMP-Hand~\citep{kuang2026jump}, as well as a mask-based visual hull baseline~\citep{scharr2017fast}. For the learning-based baselines, we follow the dataset-specific training and test splits established by POEM and POEMv2. In contrast, \ac{method} is trained exclusively on synthetic data and directly evaluated on the corresponding held-out test splits, without any fine-tuning on real-world data. Details are provided in \cref{app:sec:datasets}.

\subsection{Synthetic-to-Real Qualitative Results}\label{sec:qualitative_results}
\begin{figure}[!t]
    \centering
    \includegraphics[width=\linewidth]{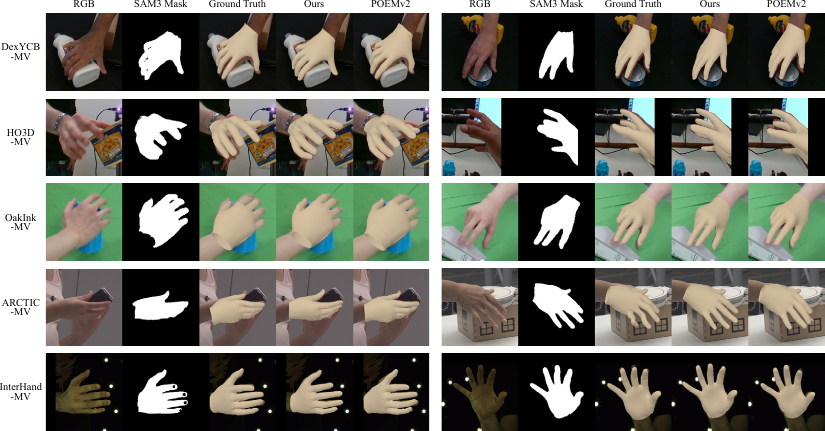}
    \caption{\textbf{Qualitative results of \ac{method}.} Using the SAM3 mask, \ac{method} exhibits better image-plane alignment than the provided 3D annotations, highlighting the annotation issue in real data.}
    \label{fig:qualitative_results}
\end{figure}

We first present the qualitative comparison of \ac{method} against the open-source POEM series~\citep{yang2023poem,yang2025multi} and the dataset ground-truth annotations in \cref{fig:qualitative_results}. The required hand segmentation masks are obtained using SAM3~\citep{carion2026sam} with the text prompt ``right hand''. We project the reconstructed hands from \ac{method} and POEMv2, as well as the dataset-annotated hand, onto the image plane using the corresponding camera geometry. The visual accuracy can then be assessed by examining the alignment between the projected hand regions and the actual right-hand regions in the RGB images. As shown in \cref{fig:qualitative_results}, \ac{method} can achieve more accurate alignment than the annotations. This observation reveals the inherent challenges in hand annotations due to the highly complex manifold of human hands. More results using SAM3 masks are in \cref{sec:app:more_results}.

\subsection{Synthetic-to-Real Results}\label{sec:quantitative_results}

The annotation discrepancies identified in \cref{sec:qualitative_results} can create a mismatch between real-image observations and the annotated hand geometry. Consequently, a reconstruction that agrees with the captured images may still be against the annotations. Such annotation biases and errors are not captured by \ac{method}, as it is trained on aligned synthetic data. To distinguish geometric reconstruction accuracy from agreement with real observations, we adopt two complementary evaluation protocols: \emph{annotation-oriented} evaluation, which measures reconstruction accuracy using annotation-consistent inputs, and \emph{observation-oriented} evaluation, which assesses consistency with independent image evidence without relying on the hand annotations.

\paragraph{Annotation-oriented evaluation.}
We render input masks from the annotated hand geometry while preserving object-induced occlusions. This removes the mismatch between the input masks and the reference geometry, allowing us to assess \ac{method}'s reconstruction performance from multi-view masks using the conventional evaluation protocol established by~\citet{yang2023poem,yang2025multi}. Across all five datasets, we report \ac{mpvpe} and \ac{mpjpe}, together with their root-relative (RR-V/RR-J) and Procrustes-aligned (PA-V/PA-J) variants. Metric details are provided in \cref{app:sec:metrics}.

\begin{wraptable}{r}{0.4\linewidth}
\vspace{-1em}
\small

\centering
\caption{\textbf{Held-out-view mask \acs{iou}$\uparrow$.}}
\label{tab:iou_results}

\begin{tabular}{@{}cccc@{}} 
\toprule
Dataset & POEMv2 & GT & Ours \\ 
\midrule
ARCTIC-MV & 79.0 & 80.1 & 81.9 \\
InterHand-MV & 81.8 & 83.9 & 83.1 \\
\bottomrule
\end{tabular}

\vspace{-1em}
\end{wraptable}

\Cref{tab:main_results} compares \ac{method} with state-of-the-art RGB-based calibrated multi-view methods---POEM~\citep{yang2023poem}, MLPHand~\citep{yang2024mlphand}, POEMv2~\citep{yang2025multi}, and JUMP-Hand~\citep{kuang2026jump}---and a mask-based visual-hull baseline~\citep{scharr2017fast} (detailed in \cref{app:sec:visual_hull_baseline}). Under this annotation-oriented protocol, \ac{method} reduces \ac{mpvpe} by \qtyrange{49.2}{63.4}{\%} across five datasets. Gains persist under root-relative and Procrustes alignment, indicating improved articulation beyond global hand placement.

\paragraph{Observation-oriented evaluation.}
To assess \ac{method}'s reconstruction fidelity on real observations independently of dataset-provided 3D hand annotations, we evaluate its agreement with hand masks from a held-out camera view on ARCTIC-MV and InterHand-MV. These are the only two datasets with enough viewpoints to reserve an additional camera exclusively for evaluation. In both experiments, \ac{method} takes as input masks extracted from real images rather than masks rendered from annotated geometry. The full evaluation protocol is described in \cref{app:sec:held_out_evaluation}.

\begin{wraptable}{r}{0.51\linewidth}
\vspace{-1em}
\small

\centering
\caption{\textbf{Synthetic mask corruption levels.}}
\label{tab:mask_corruption_calibration}

\begin{tabular}{@{}cccccc@{}} 
\toprule
\multirow{2}{*}{Metric} & \multicolumn{4}{c}{Synthetic Corruption Preset} & \multirow{2}{*}{SAM3} \\ 
\cmidrule{2-5}
 & C1 & C2 & C3 & C4 & \\ 
\midrule
\acs{iou} $\uparrow$ & 0.97 & 0.93 & 0.89 & 0.86 & 0.90 \\
Boundary F1 $\uparrow$ & 0.90 & 0.86 & 0.81 & 0.76 & 0.84 \\
\bottomrule
\end{tabular}

\vspace{-0.5em}
\end{wraptable}

We report held-out view \ac{iou} between the projected and reference hand masks, excluding visible object pixels to avoid penalizing object-occluded hand regions. As shown in \cref{tab:iou_results}, \ac{method} outperforms the baseline on both datasets, exceeding annotation \ac{iou} on ARCTIC-MV and approaching it on InterHand-MV. This advantage provides complementary evidence of its transfer to real-image observations. Additional comparisons based on 3D reconstruction metrics are provided in \cref{app:evaluation_palm}, while ablation studies investigating the effective use of mask information are presented in \cref{app:sec:ablation}.

\subsection{Scaling Synthetic Geometry Supervision}

\begin{wrapfigure}{r}{0.26652\linewidth}
    \vspace{-3.5em}

    \centering
    \includegraphics[width=\linewidth]{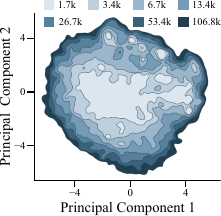}
    \caption{Geometry diversity as training size grows.}
    \label{fig:pose_param_distribution}

    \vspace{-2em}
\end{wrapfigure}

\ac{method} benefits from scalable synthetic supervision. \Cref{fig:pose_param_distribution} visualizes the generated hand geometries in the space of the first two principal components. Increasing the training set from 1.7k to 106.8k samples progressively broadens its coverage, primarily through the inclusion of more diverse hand poses. Meanwhile, reconstruction error steadily decreases as the training set grows (\cref{fig:overview}(b)). Together, these results highlight the benefits of scaling synthetic supervision in both geometric coverage and reconstruction accuracy, without the acquisition and annotation burden of expanding real-capture datasets. 

\subsection{Robustness to Imperfect Masks}

\begin{wrapfigure}{r}{0.3\linewidth}
    \vspace{-1em}

    \centering
    \includegraphics[width=\linewidth]{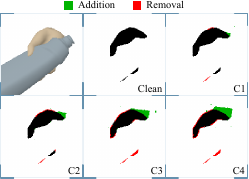}
    \caption{Illustration of mask corruption levels.}
    \label{fig:mask_corruption_example}

    \vspace{-0.5em}    
\end{wrapfigure}

While clean masks can be readily obtained through manual annotation~\citep{lin2014microsoft,he2017mask,carion2026sam}, we further investigate the robustness of \ac{method} to imperfect annotations. To ensure that the simulated annotation noise remains within a reasonable range, we use the performance of the state-of-the-art SAM3 as a calibration target. Specifically, we calibrate five mask-corruption operators to reproduce different levels of mask quality. The calibration results are reported in \cref{tab:mask_corruption_calibration}, with the corresponding evaluation protocol detailed in \cref{app:sec:corruption_calibration}. Based on these results, we define four corruption severity levels, C1--C4, with increasing degrees of mask degradation, where C3 most closely matches the performance of SAM3 (\cref{tab:mask_corruption_calibration}). Representative examples of the resulting corrupted masks are shown in \cref{fig:mask_corruption_example}.

The robustness results are presented in \cref{tab:main_results}. \ac{method} remains robust across all corruption levels and consistently retains the hand reconstruction accuracy. Even under the most severe corruption, the reconstruction metric degrades by at most \qty{1.63}{\milli\metre}. These results demonstrate that \ac{method} is robust to segmentation noise and does not rely on highly accurate mask annotations to maintain its reconstruction performance.

\section{Discussion and Limitations}

\begin{figure}[!t]
    \centering
    \includegraphics[width=\linewidth]{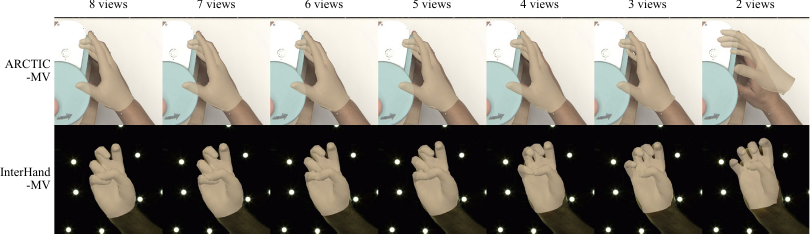}
    \caption{\textbf{Qualitative ablation of the number of input views.} \ac{method} performance is shown with the number of input views decreasing from eight to two. All input masks are predicted by SAM3.}
    \label{app:fig:view_ablation}
\end{figure}

Reconstruction accuracy naturally degrades as input views are removed (\cref{app:fig:view_ablation,app:sec:sensitivity_camera}), since fewer observations provide weaker geometric constraints on the underlying hand configuration (\cref{eq:info_structure}). We include this ablation to characterize the operating range of \ac{method} as the number of available views decreases. Moreover, since \ac{method} relies on projecting 3D grid points into 2D views to query and aggregate mask and ray features, it depends on accurate camera parameters. Errors in camera calibration can lead to inaccurate projections and misaligned cross-view evidence and degrade reconstruction accuracy (\cref{app:sec:sensitivity_camera}). Improving robustness to calibration uncertainty would therefore be an important direction for extending the method to less controlled capture settings.

Beyond limited-view conditions, an inherent limitation also arises from the ambiguity of binary masks. Although masks preserve important geometric information, they discard appearance cues such as color, texture, and local visual patterns that could help distinguish different hand configurations. Consequently, distinct 3D hand geometries may produce similar 2D masks, particularly in regions affected by self-occlusion or hand-object occlusion. While calibrated observations from multiple viewpoints help reduce this ambiguity (\cref{eq:info_structure}), they cannot always uniquely determine the underlying geometry, especially when informative contours are missing across views (\cref{app:evaluation_palm}).

A promising direction for future work is to incorporate complementary appearance cues into the reconstruction framework. For example, combining mask-based geometric evidence with RGB features or local appearance descriptors could provide additional information for resolving ambiguous hand configurations, while retaining binary masks as a robust geometric representation. An important challenge, however, is to leverage such appearance information without reintroducing the substantial synthetic-to-real domain gap that our mask-centric formulation is designed to avoid. Improving robustness under limited-view conditions, resolving mask-induced ambiguities, and exploring the optimal composition of synthetic geometric supervision would further extend the capabilities of \ac{method}.

\section{Conclusion}
We introduced \ac{method}, a calibrated multi-view hand reconstruction framework that investigates binary masks as an effective interface for bridging synthetic geometry and real observations. Results across five real-world benchmarks demonstrate the effectiveness of this mask-centric representation for synthetic-to-real transfer without photorealistic rendering or real-data fine-tuning. Further experiments highlight the benefits of scaling synthetic geometric diversity and the robustness of the approach to imperfect masks, suggesting a promising direction for scalable hand reconstruction.

\clearpage

\subsection*{AI use statement}

In this work, we used generative AI tools for language polishing and code review.

\bibliography{reference_header,reference}
\bibliographystyle{iclr2027_conference}

\clearpage
\appendix
\section{Appendix}

\subsection{Evaluation Protocol}\label{app:sec:protocol}

\subsubsection{Datasets}\label{app:sec:datasets}

We evaluate \acs{method} on five real-world benchmarks spanning single-hand motion, interacting-hand motion, and hand-object manipulation. We follow the calibrated multi-view protocols of POEM and POEMv2~\citep{yang2023poem,yang2025multi}, with right hands as reconstruction targets.

\paragraph{HO3D-MV.}
HO3D~\citep{hampali2020honnotate} captures hand-object interactions with substantial occlusion. POEM constructs HO3D-MV from sequences in HO3D v3 with complete observations from five synchronized cameras. We evaluate on the designated test sequences, GPMF1 and SB1.

\paragraph{InterHand-MV.}
InterHand2.6M~\citep{moon2020interhand} captures both single-hand and interacting-hand motions using a dense calibrated camera array. POEMv2 organizes the synchronized observations into groups of eight views to construct InterHand-MV. We evaluate on the resulting multi-view subset of the official InterHand2.6M validation set, following POEMv2.

\paragraph{DexYCB-MV.}
DexYCB~\citep{chao2021dexycb} records ten participants grasping YCB~\citep{calli2015ycb} objects. POEM constructs DexYCB-MV by grouping synchronized right-hand observations from eight cameras while retaining the official S0 split. We evaluate on the corresponding test subset.

\paragraph{OakInk-MV.}
OakInk~\citep{yang2022oakink} captures hand-object manipulation and interpersonal object handovers. POEM constructs four-view samples for each target hand, treating the hands in handover scenes separately. We evaluate on the test subset of the SP2 split with one-in-four sampling, following POEM.

\paragraph{ARCTIC-MV.}
ARCTIC~\citep{fan2023arctic} captures bimanual manipulation of articulated objects with eight static cameras and one egocentric camera. POEMv2 constructs ARCTIC-MV from synchronized observations of the right hand across the static cameras. We evaluate on the P1 validation subsets, following POEMv2.

\paragraph{Held-out-view evaluation subsets.}
We construct held-out-view subsets of ARCTIC-MV and InterHand-MV for observation-oriented evaluation (\cref{sec:quantitative_results}). On ARCTIC-MV, the eight static cameras provide inputs, while the egocentric view is reserved for evaluation. On InterHand-MV, we randomly select four existing eight-view groups per timestamp and uniformly sample an additional camera outside each group's inputs for evaluation.

\subsubsection{Metrics}\label{app:sec:metrics}

We evaluate the vertices $\hat{\mathcal{V}}$ and joints $\hat{\mathcal{J}}$ reconstructed from $\hat{\mathcal{H}}$ against the reference geometry $\mathcal{V}^*$ and $\mathcal{J}^*$ using \acf{mpvpe} and \acf{mpjpe}, respectively. With world-space coordinates in meters, these errors are computed in millimeters as
\begin{equation}
\begin{aligned}
    \operatorname{MPVPE} &= \frac{1000}{N_{\mathcal{V}}}\sum_{i=1}^{N_{\mathcal{V}}}\|\hat{\boldsymbol{v}}_i-\boldsymbol{v}_i^*\|_2,\\
    \operatorname{MPJPE} &= \frac{1000}{N_{\mathcal{J}}}\sum_{i=1}^{N_{\mathcal{J}}}\|\hat{\boldsymbol{j}}_i-\boldsymbol{j}_i^*\|_2.
\end{aligned}
\end{equation}
Root-relative errors, RR-V and RR-J, subtract the corresponding wrist position from each point set. Procrustes-aligned errors, PA-V and PA-J, additionally remove global scale and rigid transformation through a similarity transformation.

The comparison in \cref{fig:overview}(a) summarizes performance using the ratio of the lowest \ac{mpvpe} among the compared methods to each method's \ac{mpvpe} on the same dataset. Higher scores indicate better performance, with a score of one denoting the best.

\subsection{Implementation Details of \acs{method}}

\subsubsection{Robust Mask-Centric Workspace Estimation}\label{app:sec:metric_grid}

The mask-centric workspace provides a shared metric domain for aggregating multi-view observations. We construct it by estimating its center $\boldsymbol{c}$ from the observed binary masks and discretizing the surrounding region at a fixed extent and resolution. Since occlusion and segmentation errors can shift individual binary mask centroids, we estimate the center from a consensus of rays originating at the camera centers and passing through the corresponding mask centroids in the image planes.

Let $E_n=(R_n, t_n)$, where $R_n\in\mathrm{SO}(3)$ and $t_n\in\mathbb{R}^3$, map world-space coordinates to camera space, and let $\bar{\boldsymbol{u}}_n$ be the homogeneous foreground centroid of the $n$-th nonempty mask. The camera center and the corresponding unit ray direction are
\begin{equation}
    \boldsymbol{o}_n = -R_n^{\mathsf{T}}\boldsymbol{t}_n, \quad \boldsymbol{d}_n = \frac{R_n^{\mathsf{T}}K_n^{-1}\bar{\boldsymbol{u}}_n}{\left\|R_n^{\mathsf{T}}K_n^{-1}\bar{\boldsymbol{u}}_n\right\|_2}.
\end{equation}
Each nonparallel ray pair whose closest points lie in front of both cameras proposes the midpoint of those closest points. We select the proposal $\boldsymbol{c}_0$ with the smallest median Euclidean distance to the rays. Its inlier set $\mathcal{I}$ includes the proposing pair and rays whose distance to $\boldsymbol{c}_0$ does not exceed the inlier tolerance.

We refine $\boldsymbol{c}_0$ by minimizing the squared perpendicular distances to the inlier lines. Denoting $A_n=I-\boldsymbol{d}_n\boldsymbol{d}_n^{\mathsf{T}}$, the objective is
\begin{equation}
    \boldsymbol{c} = \argmin_{\boldsymbol{c}\in\mathbb{R}^3}{\sum_{n\in\mathcal{I}}{\left\|A_n\left(\boldsymbol{c} - \boldsymbol{o}_n\right)\right\|_2^2} + \epsilon \left\|\boldsymbol{c} - \boldsymbol{c}_0\right\|_2^2
    },
\end{equation}
where $\epsilon=10^{-6}$ regularizes the estimate toward $\boldsymbol{c}_0$. The closed-form solution is
\begin{equation}
    \boldsymbol{c} = \left[\sum_{n\in\mathcal{I}}{A_n + \epsilon I}\right]^{-1}\left[\sum_{n\in\mathcal{I}}{A_n\boldsymbol{o}_n + \epsilon\boldsymbol{c}_0}\right].
\end{equation}
The mask-centric frame is centered at $\boldsymbol{c}$ and shares the axis orientation of the world frame. The Pl\"ucker coordinates~\citep{plucker1828analytisch} of a pixel ray in this frame are
\begin{equation}
    \boldsymbol{r}_n(\boldsymbol{u}) = \left[\boldsymbol{d}_n(\boldsymbol{u})^{\mathsf{T}}, \left((\boldsymbol{o}_n - \boldsymbol{c})\times \boldsymbol{d}_n(\boldsymbol{u})\right)^\mathsf{T}\right]^\mathsf{T},
\end{equation}
where $\boldsymbol{d}_n(\boldsymbol{u})$ is the unit direction through pixel $\boldsymbol{u}$. This representation removes dependence on the choice of world origin.

\begin{table}[!t]
\centering
\caption{\textbf{Workspace extent ablation.} Each pair reports coverage under clean masks and $C_4$ corruption, respectively. $\uparrow$ indicates higher is better.}
\label{app:tab:workspace_coverage}

\begin{tabular}{cccc}
\toprule
$L$ (m) & Grid spacing (mm) & Vertex coverage (\%) $\uparrow$ & Whole-hand coverage (\%) $\uparrow$ \\
\midrule
0.150 & 9.3750 & 88.5755 / 83.4799 & 10.98 / 5.22 \\
0.200 & 12.5000 & 99.1255 / 98.1773 & 78.78 / 65.72 \\
0.225 & 14.0625 & 99.8931 / 99.7016 & 94.28 / 89.36 \\
0.250 & 15.6250 & 99.9962 / 99.9822 & 99.78 / 98.92 \\
0.275 & 17.1875 & 99.9996 / 99.9986 & 99.98 / 99.94 \\
0.300 & 18.7500 & 100.0000 / 99.9999 & 100.00 / 99.98 \\
\bottomrule
\end{tabular}

\end{table}

\paragraph{Mask-centric workspace discretization.}
We discretize the mask-centric workspace $\mathcal{W}(\boldsymbol{c},L) = \left\{\boldsymbol{x}\in\mathbb{R}^3\mid\|\boldsymbol{x}-\boldsymbol{c}\|_{\infty} \leq L/2\right\}$ into a regular metric grid of resolution $G\times G\times G$, with spacing $\Delta=L/G$. Its grid points are
\begin{equation}
\begin{aligned}
    \boldsymbol{g}_{i,j,k} &= \left[\frac{i+0.5}{G} - 0.5, \frac{j+0.5}{G} - 0.5, \frac{k+0.5}{G} - 0.5,\right]^{\mathsf{T}},\\
    \boldsymbol{x}_{i,j,k} &= \boldsymbol{c} + L\boldsymbol{g}_{i,j,k}, \quad i,j,k\in\{0,\dots,G-1\}.
\end{aligned}
\end{equation}
Each grid point is projected into the calibrated camera views to query mask and ray features. The fused features form a shared metric 3D representation over $K=G^3$ grid points. We use a workspace extent of $L=\qty{0.25}{\metre}$ and a grid resolution of $G=16$ along each axis to balance geometric coverage and spatial resolution.

\paragraph{Workspace coverage ablation.}
We evaluate whether the chosen extent accommodates both hand geometry and errors in the mask-derived center. We randomly sample \num{1000} hand configurations from each of HO3D-MV, InterHand-MV, DexYCB-MV, OakInk-MV, and ARCTIC-MV, yielding 5,000 samples in total. Each sample is evaluated under both clean masks and the strongest corruption level $C_4$. Masks are rendered with scene-object occlusion, and the center is re-estimated for each corruption condition. We vary $L$ while fixing $G=16$. Vertex coverage measures the fraction of hand mesh vertices contained in $\mathcal{W}(\boldsymbol{c}, L)$, including occluded vertices. Whole-hand coverage measures the fraction of configurations whose vertices are all contained.

\Cref{app:tab:workspace_coverage} shows that the $L=\qty{0.25}{\metre}$ workspace retains \qty{99.9822}{\%} vertex coverage and \qty{98.92}{\%} whole-hand coverage under $C_4$, compared with \qty{99.9962}{\%} and \qty{99.78}{\%} for clean masks. Reducing $L$ to \qty{0.225}{\metre} lowers whole-hand coverage under $C_4$ to \qty{89.36}{\%}, while larger extents increase the grid spacing at fixed resolution. These results support \qty{0.25}{\metre} as a practical balance between hand coverage and spatial resolution geometry, including under severe mask corruption.

\subsubsection{Global Hand Pose Estimation from Palm Joints}
\label{app:sec:global_pose}

\begin{table}[!t]
\centering
\caption{\textbf{Global rotation error with mean-shape and reference palm templates.} Errors (degrees) are averaged over \num{5000} samples per mask condition.}
\label{app:tab:palm_sensitivity}

\begin{tabular}{ccc}
\toprule
Mask condition & Mean-shape template & Reference template \\
\midrule
Clean & 1.64 & 1.54 \\
$C_1$ & 1.79 & 1.70 \\
$C_2$ & 1.95 & 1.86 \\
$C_3$ & 2.13 & 2.05 \\
$C_4$ & 2.41 & 2.33 \\
\bottomrule
\end{tabular}

\end{table}

\begin{wrapfigure}{r}{0.3\linewidth}
    \vspace{-1em}

    \centering
    \includegraphics[width=\linewidth]{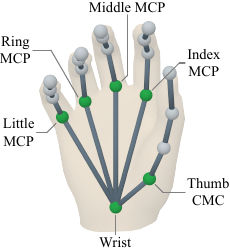}
    \caption{MANO palm joints.}
    \label{app:fig:palm_joints}
\end{wrapfigure}

We predict auxiliary world-space hand joints $\tilde{\mathcal{J}}$ from the shared metric 3D representation $\mathcal{Z}$ as geometric anchors for global hand pose estimation. In the MANO hand model, the canonical positions of the wrist and five finger-base joints are independent of the articulation parameters $\boldsymbol{\theta}$ for a fixed hand shape, providing a rigid configuration for palm alignment; see \cref{app:fig:palm_joints}. For the alignment in \cref{eq:palm_alignment}, we approximate individual palm geometry with a fixed mean-shape MANO template $\mathcal{J}_{\text{palm}}^c$ obtained from $h(\boldsymbol{0},\boldsymbol{0})$. The template determines the estimated rotation; translation is given directly by the predicted wrist. This permits global pose estimation before articulation and shape decoding.

\paragraph{Effect of palm-shape variation.}
We evaluate the effect of neglecting individual palm-shape variation on global hand rotation estimation by comparing the fixed mean-shape template with a reference template obtained from $h(\boldsymbol{0},\boldsymbol{\beta}^*)$. We randomly draw \num{1000} samples from each of HO3D-MV, InterHand-MV, DexYCB-MV, OakInk-MV, and ARCTIC-MV. Each sample is evaluated under clean masks and the four corruption levels $C_1$--$C_4$, with scene-object occlusion retained. Both templates are aligned to the same predicted palm joints. We measure global rotation error as the angular distance between the predicted and ground-truth rotations, expressed in degrees and averaged over the \num{5000} samples for each mask condition. \Cref{app:tab:palm_sensitivity} shows that the fixed mean-shape template increases mean rotation error by at most \qty{0.1}{\degree} relative to the reference template across all five mask conditions. These results support using a fixed-mean-shape palm template for global rotation estimation under both clean and corrupted masks.

\subsubsection{Hand Shape Parameter Recovery by Projection}\label{app:sec:shape_projection}

Shape projection converts the predicted neutral canonical hand geometry into valid MANO shape parameters without a learned parameter-regression head. Let $\mathcal{V}^0(\boldsymbol{\beta})$ denote the vertex output of $h(\boldsymbol{0},\boldsymbol{\beta})$ in \cref{eq:hand_model}, with articulation fixed to zero. The shape token predicts displacements from the mean-shape geometry $\mathcal{V}^0(\boldsymbol{0})$, yielding the neutral canonical geometry $\hat{\mathcal{V}}^0$.

At zero articulation, the hand model depends affinely on $\boldsymbol{\beta}$~\citep{romero2017embodied}, and vertex displacements relative to $\mathcal{V}^{0}(\boldsymbol{0})$ therefore lie in a fixed linear shape subspace. To make shape parameter recovery insensitive to translation errors in the predicted geometry, we subtract the vertex centroid. For the hand mesh vertices $\mathcal{V}$, its centered vertices are
\begin{equation}
    \overline{\boldsymbol{v}}_i=\boldsymbol{v}_i- \frac{1}{N_\mathcal{V}}\sum_{j=1}^{N_\mathcal{V}}\boldsymbol{v}_j, \quad \overline{\mathcal{V}}=\{\overline{\boldsymbol{v}}_i\}_{i=1}^{N_\mathcal{V}}.
\end{equation}
The fixed shape basis $S\in\mathbb{R}^{3N_{\mathcal{V}}\times N_{\boldsymbol{\beta}}}$ has columns $\operatorname{vec}\left(\overline{\mathcal{V}^0}(\boldsymbol{e}_r) - \overline{\mathcal{V}^0}(\boldsymbol{0})\right)$, where $\operatorname{vec}(\cdot)$ denotes the matrix vectorization and $\boldsymbol{e}_r$ is the $r$-th unit vector. Since the predicted displacements are not constrained to the shape subspace, we recover the shape parameters through least-squares projection onto this subspace:
\begin{equation}
\begin{aligned}
    \hat{\boldsymbol{\beta}} &= \argmin_{\boldsymbol{\beta}\in\mathbb{R}^{N_{\boldsymbol{\beta}}}} \left\| S\boldsymbol{\beta} -\operatorname{vec}\left(\overline{\hat{\mathcal{V}}^0} - \overline{\mathcal{V}^0}(\boldsymbol{0})\right) \right\|_2^2 \\
    &=S^{+}\operatorname{vec}\left(\overline{\hat{\mathcal{V}}^{0}}-\overline{\mathcal{V}^{0}}(\boldsymbol{0})\right),
\end{aligned}
\end{equation}
where $S^{+}$ is the Moore-Penrose pseudoinverse of $S$. Both $S$ and $S^{+}$ are fixed by the hand model and can be precomputed. Projection retains the component of the predicted displacements within the MANO shape subspace, combining geometric prediction with a compact parametric output.

\begin{table}[!t]
\centering
\caption{\textbf{Neutral-geometry accuracy before and after shape projection.} \acs{mpvpe} is measured against the reference neutral geometry and averaged over \num{5000} samples per mask condition. $\Delta$ denotes the change after projection (after minus before).}
\label{app:tab:shape_projection}

\begin{tabular}{cccc}
\toprule
Mask condition & Before projection & After projection & $\Delta$ \\
\midrule
Clean & 1.0250 & 1.0251 & +0.000099 \\
$C_1$ & 1.0955 & 1.0956 & +0.000064 \\
$C_2$ & 1.1473 & 1.1474 & +0.000093 \\
$C_3$ & 1.2008 & 1.2010 & +0.000127 \\
$C_4$ & 1.2781 & 1.2782 & +0.000155 \\
\bottomrule
\end{tabular}

\end{table}

\paragraph{Reconstruction accuracy.}
We evaluate whether projection degrades reconstruction accuracy using \num{1000} samples from each of HO3D-MV, InterHand-MV, DexYCB-MV, OakInk-MV, and ARCTIC-MV. Each sample is evaluated under clean masks and the four corruption levels $C_1$--$C_4$, with scene-object occlusion retained. \ac{mpvpe} before and after projection is measured against the same reference neutral geometry $\mathcal{V}^{0,*}=\mathcal{V}^0(\boldsymbol{\beta}^*)$. \Cref{app:tab:shape_projection} shows that the projection increases mean \ac{mpvpe} by at most \qty{0.000155}{\milli\metre} across all five conditions, indicating that shape parameters can be recovered with negligible loss of reconstruction accuracy.

\subsubsection{Network Architecture}

\ac{method} aggregates multi-view masks into a shared metric 3D representation for global hand pose estimation and canonical hand state decoding. We detail the network architecture in \cref{app:tab:network_architecture}.

\begin{table}[!t]
\centering
\caption{\textbf{Network architecture of \acs{method}.}}
\label{app:tab:network_architecture}

\begin{tabular}{p{0.22\linewidth}>{\raggedright\arraybackslash}m{0.72\linewidth}}
\toprule
Component & Configuration \\
\midrule
Input masks & Resolution of $224\times224$ pixels per view. \\
\midrule
Mask encoder & Four-stage ConvNeXt-V2-style feature pyramid~\citep{woo2023convnext} with stage widths $(24,48,96,192)$, depths $(1,1,2,2)$, and FFN expansion ratio 4. The stem and inter-stage downsampling use strides 4 and 2, respectively. \\
Ray encoder & A $4\times4$ convolution with stride 4, followed by a residual block with $3\times3$ depthwise and $1\times1$ convolutions. \\
\midrule
Metric grid & $16\times16\times16$ grid points discretizing a workspace of extent \qty{0.25}{\metre}. \\
Positional encoders & Coordinates concatenated with Fourier features at frequencies $(1,2,4,8)$, followed by two-layer MLPs of width 64 for cross-view aggregation and 128 for canonical hand state decoding. \\
Cross-view attention & Two attention blocks with feature width 64, 4 heads, and FFN expansion ratio 4. \\
3D CNN & Three residual blocks with feature width 64, $3\times3\times3$ depthwise convolutions, layer normalization, global response normalization, and FFN expansion ratio 2. \\
\midrule
Joint head & Two $1\times1\times1$ convolutions predicting $N_{\mathcal{J}}$ spatial logits and $3N_{\mathcal{J}}$ offset components at each grid point. \\
Token-query decoder & $N_{\boldsymbol{\theta}}$ pose tokens and one shape token processed by three decoder blocks, each comprising self-attention, cross-attention, and an FFN. Feature width 128, 8 attention heads, and FFN expansion ratio 4. \\
Articulation head & A shared two-layer MLP with layer normalization, GELU activation, and hidden width 128, producing one scalar per pose token. \\
Shape head & A two-layer MLP with layer normalization, GELU activation, and hidden width 128, producing $3N_{\mathcal{V}}$ vertex displacements. \\
\midrule
Trainable parameters & Approximately 2.31 million. \\
\bottomrule
\end{tabular}
\end{table}

\subsection{Training Details}\label{app:sec:training_details}

\paragraph{Optimization.}
We train \ac{method} from scratch for \num{1000000} steps using AdamW with batch size 32, peak learning rate \num{0.001}, weight decay \num{0.01}, and moment coefficients $(0.9,0.95)$. The learning rate increases to its peak over the first \num{5000} steps (\qty{0.5}{\%} of training) and then follows a cosine decay schedule. We compute the vertex and joint smooth-$\ell_1$ losses with residuals expressed in centimeters and set the loss weights in \cref{eq:total_loss} to $(\lambda_{\text{pose}},\lambda_{\text{shape}},\lambda_{\text{joint}},\lambda_{\text{rot}})=(1,1,1,0.01)$. All experiments are trained on a single NVIDIA RTX 5090 and evaluated using the final checkpoint. The experiments in \cref{fig:overview}(b) all use the same training budget of \num{1000000} optimization steps.

\begin{table}[!t]
\centering
\caption{\textbf{Training observation augmentations.} $p_s$ and $p_v$ denote the per-sample and conditional per-view probabilities, respectively.}
\label{app:tab:augmentation}

\begin{tabular}{>{\raggedright\arraybackslash}m{0.30\linewidth}cc>{\raggedright\arraybackslash}m{0.51\linewidth}}
\toprule
Augmentation & $p_s$ & $p_v$ & Description \\
\midrule
Hand-region visibility
& 0.5 & 1.0
& Occlusion by non-target geometry and partial hand-region removal. \\
\midrule
Forearm extent
& 0.5 & 0.8
& Extending the forearm with length scale $[0,0.6]$. \\
\midrule
Morphological perturbations
& 0.5 & 0.8
& Filling small gaps in the mask using a diamond-shaped kernel with radius $[1,4]$ pixels. \\
\midrule
Boundary perturbations
& 0.5 & 0.8
& Smoothly displacing the mask boundary with amplitude in
$[0.55,\max(0.55,0.0025\min(H,W))]$ pixels. \\
\midrule
Localized boundary artifacts
& 0.5 & 0.8
& Adding $[1,4]$ square artifacts near the boundary, with side lengths $[1,3]$ pixels. \\
\midrule
Random view dropout
& 0.5 & 0.2
& Randomly dropping input views while retaining at least three. \\
\bottomrule
\end{tabular}

\end{table}

\paragraph{Augmentation.}
\Cref{app:tab:augmentation} specifies the training augmentations. Here $p_s$ is the probability of enabling an augmentation for a multi-view sample, and $p_v$ is its conditional probability for each view. These operations alter the observations while retaining the same target hand geometry. Random view dropout retains at least three input views. The training probabilities are distinct from the fixed severity presets used for robustness evaluation in \cref{app:sec:corruption_calibration}.

\subsection{Ablation Studies}\label{app:sec:ablation}

We evaluate the contributions of cross-view attention, joint-based global-pose estimation, and canonicalization to reconstruction accuracy. The full model constructs a metric 3D representation through cross-view attention, estimates global pose from auxiliary palm joints, and decodes articulation and shape using canonical coordinates. We compare it with three variants that modify the corresponding stages.

\paragraph{Direct global-pose regression.}
This variant predicts global rotation and translation directly from the shared metric 3D representation, replacing auxiliary joint prediction and palm alignment. It replaces the auxiliary joint loss with a smooth-$\ell_1$ translation loss computed in centimeters and assigns unit weights to translation and rotation losses. The comparison evaluates the benefit of estimating global pose through explicit joint geometry.

\paragraph{Without canonicalization.}
This variant retains auxiliary joint prediction and palm alignment but conditions the token-query decoder on mask-centric coordinates instead of coordinates transformed by $\hat{T}^{-1}$. This comparison evaluates the contribution of canonical coordinates to articulation and shape decoding.

\paragraph{Mean Pooling.}
This variant replaces the cross-view attention with uniform aggregation of per-view features. At each 3D grid point, a shared MLP jointly encodes the queried mask and Pl\"ucker ray features from each view. The resulting features are mean-pooled across views before positional encoding and residual feed-forward blocks are applied. The remaining reconstruction stages and losses are unchanged. This comparison evaluates the proposed fusion module against a mean-pooling alternative that retains both input modalities.

All variants use the full synthetic training set and the optimization settings in \cref{app:sec:training_details}, with the loss modifications described above for direct global-pose regression. As shown in \cref{app:tab:ablation}, all three variants increase \ac{mpvpe} and PA-V across all five mask conditions. These results support the proposed cross-view fusion, joint-based global-pose estimation, and canonical decoding under the evaluated mask conditions.

\begin{table}[!t]
\centering
\caption{\textbf{Effect of cross-view attention, joint-based global pose estimation, and canonicalization.} Each experiment reports \acs{mpvpe} / PA-V, with the lowest errors highlighted in \textbf{bold}.}
\label{app:tab:ablation}

\begin{tabular}{@{}cccccc@{}}
\toprule
Variant & Clean & $C_1$ & $C_2$ & $C_3$ & $C_4$ \\
\midrule
\acs{method} & \textbf{2.86} / \textbf{1.92} & \textbf{3.05} / \textbf{2.02} & \textbf{3.30} / \textbf{2.15} & \textbf{3.50} / \textbf{2.26} & \textbf{3.77} / \textbf{2.39} \\
Direct global pose regression & 3.39 / 2.60 & 3.64 / 2.72 & 3.85 / 2.85 & 4.08 / 2.97 & 4.34 / 3.12 \\
Without canonicalization & 3.18 / 2.24 & 3.37 / 2.36 & 3.56 / 2.50 & 3.80 / 2.63 & 4.10 / 2.79 \\
Mean pooling & 3.01 / 2.06 & 3.22 / 2.16 & 3.48 / 2.29 & 3.71 / 2.41 & 4.03 / 2.55 \\
\bottomrule
\end{tabular}
\end{table}

\subsection{Visual Hull Baseline}\label{app:sec:visual_hull_baseline}

We evaluate a mask-based geometric baseline that reconstructs a visual hull and fits the MANO-based hand model in \cref{eq:hand_model}. The baseline receives the same calibrated $224\times224$ clean masks as \ac{method}, with scene-object occlusion retained. To favor this baseline, we provide ground-truth global hand pose $T^*$ and shape parameter $\boldsymbol{\beta}^*$, leaving only the $N_{\boldsymbol{\theta}}$ articulation parameters $\boldsymbol{\theta}$ to be estimated.

\paragraph{Visual hull reconstruction.}
We reconstruct the hull using \texttt{torchhull}, a GPU implementation of the volume-carving approach of \citet{scharr2017fast}. Reconstruction is performed within a \qty{0.25}{\metre} cube centered using the mask-based estimate in \cref{app:sec:metric_grid}, at approximately \qty{1.95}{\milli\metre} resolution. To accommodate missing foreground evidence under occlusion, we relax the all-view intersection, inspired by the soft visual-hull formulation of \citet{narayan2015range}. We use a support threshold of $k=\max(2,\lceil0.75N\rceil)$, where $N$ is the number of nonempty valid masks, and retain the standard intersection when $k=N$. The support fraction is fixed at 0.75 across datasets.

\paragraph{Articulation parameter fitting.}
We fit the hand articulation to the reconstructed hull using a robust bidirectional geometric objective. We sample \num{2048} points uniformly by surface area from the hull and transform them into the canonical hand frame using $(T^*)^{-1}$. The resulting point set $\mathcal{P}$ remains fixed throughout optimization. Let $\mathcal{M}_{\boldsymbol{\theta}}^c$ and $\mathcal{V}_{\boldsymbol{\theta}}^c$ denote the mesh and mesh vertices of the hand model $h(\boldsymbol{\theta},\boldsymbol{\beta}^*)$, respectively. We minimize
\begin{equation}
    \mathcal{L}_{\text{fit}}= \frac{1}{|\mathcal{P}|}\sum_{\boldsymbol{p}\in\mathcal{P}}\rho_{\delta}\left(d(\boldsymbol{p},\mathcal{M}_{\boldsymbol{\theta}}^c)\right) + \frac{0.25}{|\mathcal{V}_{\boldsymbol{\theta}}^c|}\sum_{\boldsymbol{v}\in\mathcal{V}_{\boldsymbol{\theta}}^c}\rho_{\delta}\left(d(\boldsymbol{v},\mathcal{P})\right)+\frac{10^{-3}}{N_{\boldsymbol{\theta}}}\|\boldsymbol{\theta}\|_2^2.
\end{equation}
The hull-to-hand term measures point-to-triangle distances to the hand mesh, while the hand-to-hull term measures nearest-point distances to the sampled hull. We express distances in centimeters and use the Huber loss as the robust penalty
\begin{equation}
    \rho_{\delta}(d)=
    \begin{cases}
        d^2, & d\leq\delta,\\
        2\delta d-\delta^2, & d>\delta,
    \end{cases}
\end{equation}
with $\delta=1$, corresponding to \qty{10}{\milli\metre}. This penalty limits the influence of large geometric discrepancies. The lower weight on the hand-to-hull term reduces the penalty on hand regions absent from the reconstruction due to occlusion, while the final term regularizes the articulation parameters toward zero.

\paragraph{Optimization and evaluation.}
We optimize articulation subject to anatomical joint limits, keeping global pose and shape fixed. We use three generic initializations: one with zero articulation offsets and two with increasing finger flexion. Each receives \num{50} Adam updates at a learning rate of \num{0.01}. We empirically verified that convergence is reached well within this optimization budget. We select the iterate with the lowest fitting objective across all runs. Ground-truth articulation is not used for initialization, optimization, or solution selection. Across the evaluation samples in \cref{app:sec:datasets}, 44 yield empty hulls and are excluded; reported errors are averaged over successful reconstructions. Results on the five real-world datasets are reported in \cref{tab:main_results}. We report only PA-V and PA-J, as the baseline estimates articulation with global pose and shape fixed to ground truth.

\subsection{Held-Out Evaluation}\label{app:sec:held_out_evaluation}

We evaluate cross-view mask agreement on ARCTIC-MV and InterHand-MV using the subsets described in \cref{app:sec:datasets}. \ac{method} takes SAM3 masks from eight static cameras as input, whereas POEMv2 receives RGB images from the same views. We reserve the egocentric view on ARCTIC-MV and one additional camera view outside each input group on InterHand-MV exclusively for evaluation, withholding these views from both models.

\begin{wraptable}{r}{0.4\linewidth}
    \centering
    \caption{\textbf{Comparison on the PALM dataset.}}
    \label{app:tab:palm}

    \small
    \begin{tabular}{lcc}
        \toprule
        Method & Surface distance $\downarrow$ & \ac{iou} $\uparrow$ \\
        \midrule
        \ac{method} & \textbf{3.109} & \textbf{83.1} \\
        POEMv2 & 3.498 & 80.6 \\
        \bottomrule
    \end{tabular}
\end{wraptable}

We render both predicted hand geometries and the original dataset MANO annotations using the calibrated held-out cameras, accounting for occlusion by annotated non-target geometry. We compare the resulting visible hand masks against SAM3 reference masks from the held-out views. We retain only samples with non-empty held-out reference masks, yielding \num{15565} samples for ARCTIC-MV and \num{19758} samples for InterHand-MV. For each sample, we compute the \ac{iou} between the rendered visible hand mask and the held-out reference SAM3 mask, and report the median \ac{iou} in \cref{tab:iou_results}.

\subsection{Evaluation on the PALM dataset}\label{app:evaluation_palm}

We evaluate \ac{method} on the PALM dataset~\citep{fan2026palm}, which provides high-precision 3D scans of hand surfaces. \ac{method} takes SAM3-generated masks as input, whereas POEMv2 operates on RGB images captured by the same cameras. We quantify reconstruction accuracy using the mean point-to-surface distance, in millimeters, from the predicted hand meshes to the ground-truth scans. By directly evaluating against the scanned surfaces, this protocol avoids potential errors introduced by fitting the MANO model to the reference geometry. As shown in \cref{app:tab:palm}, \ac{method} achieves a mean surface distance of \SI{3.109}{\milli\metre}, outperforming POEMv2 by \SI{0.389}{\milli\metre}, with consistent \ac{iou} improvement by 2.5 percentage point.

\vspace{-20pt}

\paragraph{Articulation disambiguation under similar masks.}
To better understand how the distinguishability of input masks affects \ac{method}'s ability to recover articulation changes, we conduct a controlled experiment using randomly sampled hand articulations from the synthetic test set under PALM's camera configuration. For each articulation, we perturb one finger's joint parameter in a randomly selected positive or negative direction, while keeping hand shape, global pose, other joints, and camera parameters fixed.

We consider two groups: (i) Nearly identical masks: small articulation perturbations of $5^\circ$--$10^\circ$, with mask \ac{iou} $\ge$ 0.99 in every camera view before and after perturbation. These cases represent highly ambiguous observations, where articulation changes introduce almost no distinguishable mask information. (ii) Distinguishable masks: larger articulation perturbations of $30^\circ$--$45^\circ$, with mask \ac{iou} $\le$ 0.90 in at least one camera view. These cases represent scenarios where at least one view provides additional geometric evidence of the articulation change. All configurations satisfy anatomical joint limits and self-collision constraints. We randomly select \num{5000} valid perturbed instances (\num{1000} perturbations per finger) for each group and evaluate fingertip position errors before and after Procrustes alignment (PA).

To directly measure articulation discrimination, we evaluate whether the predicted fingertip displacement between the original and perturbed poses agrees with the ground-truth displacement. Let $\mathbf{p}_0,\mathbf{p}_1$ denote the ground-truth fingertip positions before and after perturbation, and $\hat{\mathbf{p}}_0,\hat{\mathbf{p}}_1$ their corresponding predictions. We define the relative displacement-vector error as
\begin{equation}
   e = \frac{\left\|(\hat{\mathbf{p}}_1-\hat{\mathbf{p}}_0)-(\mathbf{p}_1-\mathbf{p}_0)\right\|_2}{\|\mathbf{p}_1-\mathbf{p}_0\|_2} \times 100\,\%. 
\end{equation}

\begin{table}[!t]
    \centering
    \caption{\textbf{Relative fingertip displacement-vector error $e$ under different mask conditions.} Lower value indicates a better result.}
    \label{app:tab:articulation_ambiguity}

    \begin{tabular}{lcc}
        \toprule
        \textbf{Metric} & \textbf{Nearly identical masks} & \textbf{Distinguishable masks} \\
        \midrule
        $e$ before PA (\%) & 29.70 & 7.18 \\
        $e$ after PA (\%) & 25.65 & 5.27 \\
        \bottomrule
    \end{tabular}
\end{table}

As shown in \cref{app:tab:articulation_ambiguity}, for nearly identical masks, \ac{method} exhibits a relative fingertip displacement error of \qty{29.70}{\%} (\qty{25.65}{\%} after PA), indicating difficulty in recovering articulation changes when the input masks provide insufficient discriminative information. In contrast, when at least one camera view captures distinguishable mask changes, the relative error decreases substantially to \qty{7.18}{\%} (\qty{5.27}{\%} after PA), demonstrating that \ac{method} can effectively respond to informative mask variations and recover the corresponding fingertip motion. These findings highlight both \ac{method}'s sensitivity to observable articulation changes and the inherent ambiguity of mask-only reconstruction when different articulations produce nearly identical masks.

\subsection{Mask-Corruption Calibration}\label{app:sec:corruption_calibration}

\begin{table}[!t]
\centering
\caption{\textbf{Mask corruption levels $C_1$--$C_4$.}}
\label{app:tab:corruption_operator}

\begin{tabular}{lcccc}
\toprule
Parameter & $C_1$ & $C_2$ & $C_3$ & $C_4$ \\
\midrule
Forearm length scale & 0.125 & 0.250 & 0.375 & 0.500 \\
Fragment-dropout probability & 0.05 & 0.10 & 0.25 & 0.50 \\
Morphological perturbations (pixels) & 1 & 2 & 3 & 4 \\
Boundary perturbation coefficient $\eta$ & 0 & $1/3$ & $2/3$ & 1 \\
Number of localized boundary artifacts & 1 & 2 & 3 & 4 \\
\bottomrule
\end{tabular}
\end{table}

\subsubsection{Mask-Corruption Operators}

We evaluate reconstruction robustness by applying controlled perturbations to annotation-consistent masks. The four presets $C_1$--$C_4$ in \cref{app:tab:corruption_operator} combine five operators in a fixed order: forearm extent, fragment dropout, morphological perturbations, boundary perturbations, and localized boundary artifacts. The reference hand geometry is held fixed across all corruption levels.

Forearm extension adds a forearm region at the wrist according to the prescribed length scale. Fragment dropout removes small disconnected foreground regions. For masks with multiple components, a component of area $A\leq 64$ pixels is removed with probability $\min{\left(1,p/\sqrt{A/A_{\text{max}}}\right)}$, where $A_{\text{max}}$ is the largest component area and $p$ is the dropout probability. Morphological perturbations and localized artifacts use the operations described in \cref{app:tab:augmentation}, with the severity parameters in \cref{app:tab:corruption_operator}. Boundary perturbations use displacement amplitude $0.55+\eta(a_{\text{max}}-0.55)$ pixels, where $a_{\text{max}}=\max(0.55,0.002\min(H,W))$ at the source image resolution $H\times W$.

\subsubsection{Calibration Protocol}

To provide a segmentation-based reference for corruption severity, we evaluate SAM3 on \num{4000} synthetic RGB images rendered using camera parameters from each of the five datasets, yielding \num{20000} images in total. Following \cref{sec:data_generation}, we generate hand geometries by sampling articulations in the MANO pose PCA space and selecting shape parameters from the empirical distribution. Appearance varies through DART skin textures~\citep{gao2022dart}, sampled lighting, and an extended forearm. Backgrounds are drawn from ImageNet~\citep{deng2009imagenet}.

We evaluate SAM3 segmentations against the rendered visible binary hand masks. To compare segmentation errors with the prescribed corruption levels, we also perturb each reference mask three times at each level and evaluate the corrupted masks against the same reference.

Mask quality is measured by median \ac{iou} and median boundary F1~\citep{bokhovkin2019boundary}. For a reference mask $B$ and its segmented or corrupted counterpart $\hat{B}$, $\operatorname{IoU}=|\hat{B}\cap B|/|\hat{B}\cup B|$. Boundary F1 is defined as the harmonic mean of boundary precision and recall. We extract inner boundaries using $3\times3$ erosion and match boundary pixels within a Euclidean distance of $\tau=\max(1,\operatorname{round}(0.01\sqrt{|B|}))$ pixels at the source image resolution. The comparisons are provided in \cref{tab:mask_corruption_calibration}.

\subsection{Sensitivity to Camera Configurations}\label{app:sec:sensitivity_camera}

\begin{table}[!t]
    \centering
    \caption{\textbf{Sensitivity to the number of input views.} We report MPVPE in millimeters on the five real-world datasets.}
    \label{app:tab:view_count}
    
    \begin{tabular}{lccccccc}
        \toprule
        Dataset & 2 views & 3 views & 4 views & 5 views & 6 views & 7 views & 8 views \\
        \midrule
        DexYCB-MV & 34.47 & 8.23 & 4.51 & 2.78 & 2.44 & 2.21 & 2.07 \\
        HO3D-MV & 11.23 & 4.44 & 3.49 & 2.79 & -- & -- & -- \\
        OakInk-MV & 12.50 & 3.91 & 2.95 & -- & -- & -- & -- \\
        ARCTIC-MV & 74.94 & 32.13 & 10.00 & 5.81 & 3.54 & 3.14 & 2.99 \\
        InterHand-MV & 11.11 & 5.96 & 4.62 & 4.22 & 3.95 & 3.71 & 3.52 \\
        \bottomrule
    \end{tabular}
\end{table}

\paragraph{Sensitivity to the number of input views.}
We evaluate the sensitivity of \ac{method} to the number of input views on nested subsets of available cameras using rendered masks. As shown in \cref{app:tab:view_count}, reconstruction accuracy generally improves with increasing view count, suggesting that additional cameras provide increasingly geometric constraints. These results highlight the importance of sufficient multi-view coverage for accurate reconstruction.

\paragraph{Sensitivity to calibration errors.}
We further assess the robustness of \ac{method} to camera calibration errors by jointly perturbing camera translation, rotation, focal length, and principal point while keeping the input masks and reference hand geometry fixed. We consider four perturbation levels (L1--L4) with progressively increasing magnitudes, as specified in \cref{app:tab:calibration_levels}, where L0 denotes the original, unperturbed calibration. The corresponding reconstruction errors are reported in \cref{app:tab:calibration_results}. \ac{method} exhibits relatively small performance variations under mild calibration perturbations (L1--L2), while larger perturbations lead to more accuracy degradation. These results reflect the reliance of \ac{method} on accurate camera parameters, as calibration errors introduce inconsistencies in 3D-to-2D projections and cross-view geometric alignment.

\begin{table}[!t]
    \centering
    \caption{\textbf{Camera calibration perturbation levels.}}
    \label{app:tab:calibration_levels}

    \begin{tabular}{lcccc}
        \toprule
        Perturbation & L1 & L2 & L3 & L4 \\
        \midrule
        Translation (mm) & 0.10 & 0.25 & 0.50 & 1.00 \\
        Rotation ($^\circ$) & 0.025 & 0.05 & 0.10 & 0.25 \\
        Focal length (\%) & 0.10 & 0.25 & 0.50 & 1.00 \\
        Principal point (\%) & 0.025 & 0.05 & 0.10 & 0.25 \\
        \bottomrule
    \end{tabular}
\end{table}

\begin{table}[!t]
    \centering
    \caption{\textbf{Sensitivity to camera calibration errors.} We report MPVPE in millimeters under different perturbation levels, where L0 denotes unperturbed calibration.}
    \label{app:tab:calibration_results}

    \begin{tabular}{lccccc}
        \toprule
        Dataset & L0 & L1 & L2 & L3 & L4 \\
        \midrule
        DexYCB-MV & 2.07 & 2.09 & 2.05 & 2.31 & 3.62 \\
        HO3D-MV & 2.79 & 2.79 & 2.82 & 2.90 & 3.19 \\
        OakInk-MV & 2.95 & 3.09 & 3.20 & 3.58 & 5.42 \\
        ARCTIC-MV & 2.99 & 3.06 & 3.23 & 4.20 & 8.05 \\
        InterHand-MV & 3.52 & 3.55 & 3.62 & 3.69 & 4.54 \\
        \bottomrule
    \end{tabular}
\end{table}

\subsection{Additional Qualitative Results}\label{sec:app:more_results}

In this section, we present additional qualitative results to further illustrate the effectiveness of \ac{method}. All reconstructions are obtained using SAM3 masks as the model input.

\begin{figure}[!t]
    \centering
    \includegraphics[width=\linewidth]{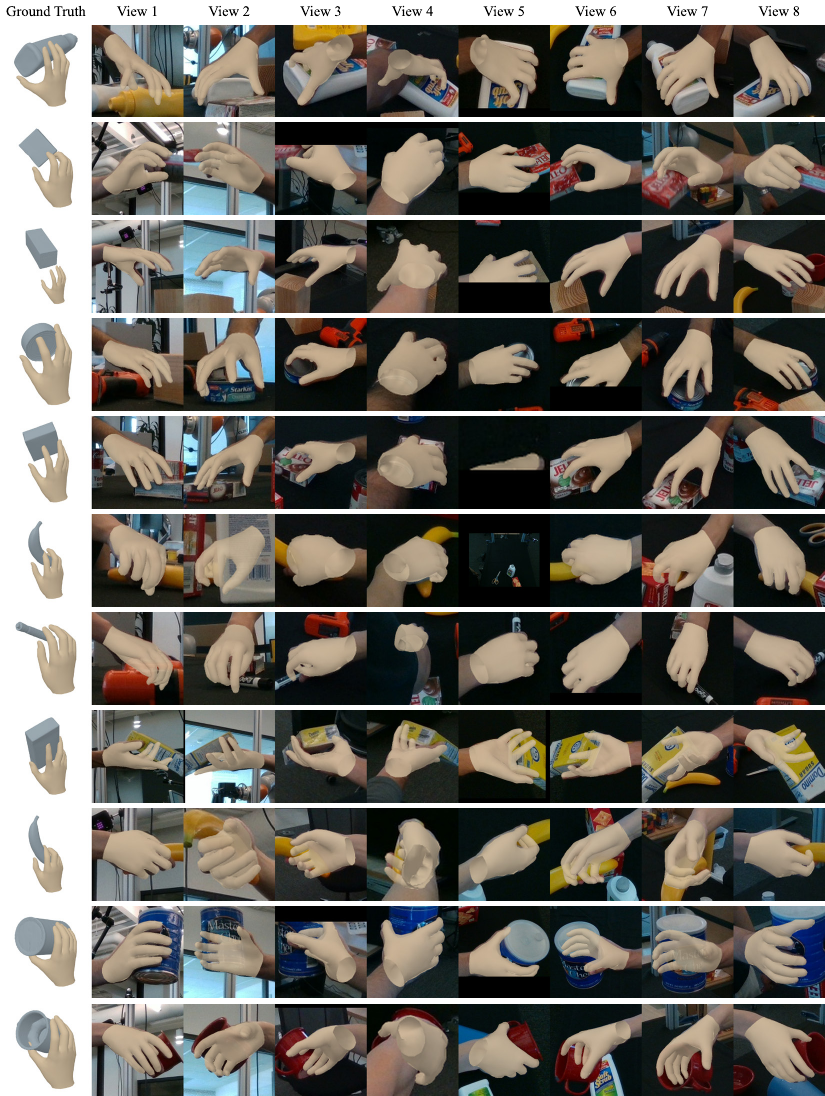}
    \caption{\textbf{Qualitative results on DexYCB-MV} with SAM3-predicted masks as input.}
\end{figure}

\begin{figure}[!t]
    \centering
    \includegraphics[width=\linewidth]{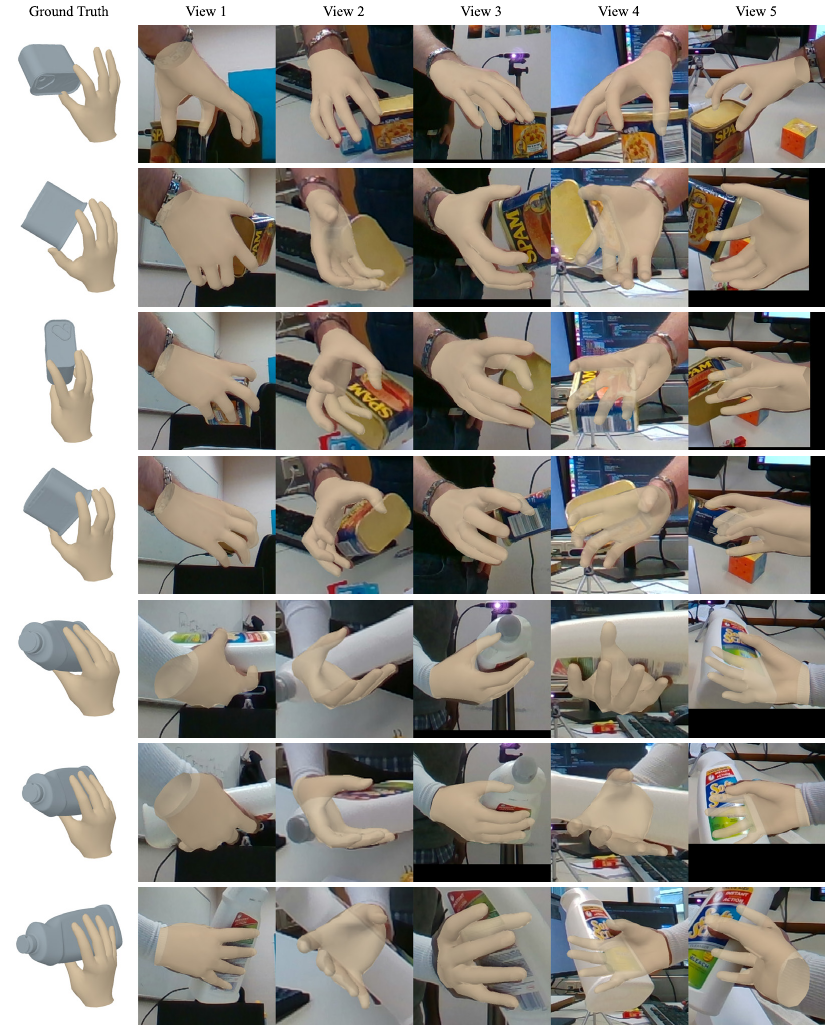}
    \caption{\textbf{Qualitative results on HO3D-MV} with SAM3-predicted masks as input.}
\end{figure}

\begin{figure}[!t]
    \centering
    \includegraphics[width=\linewidth]{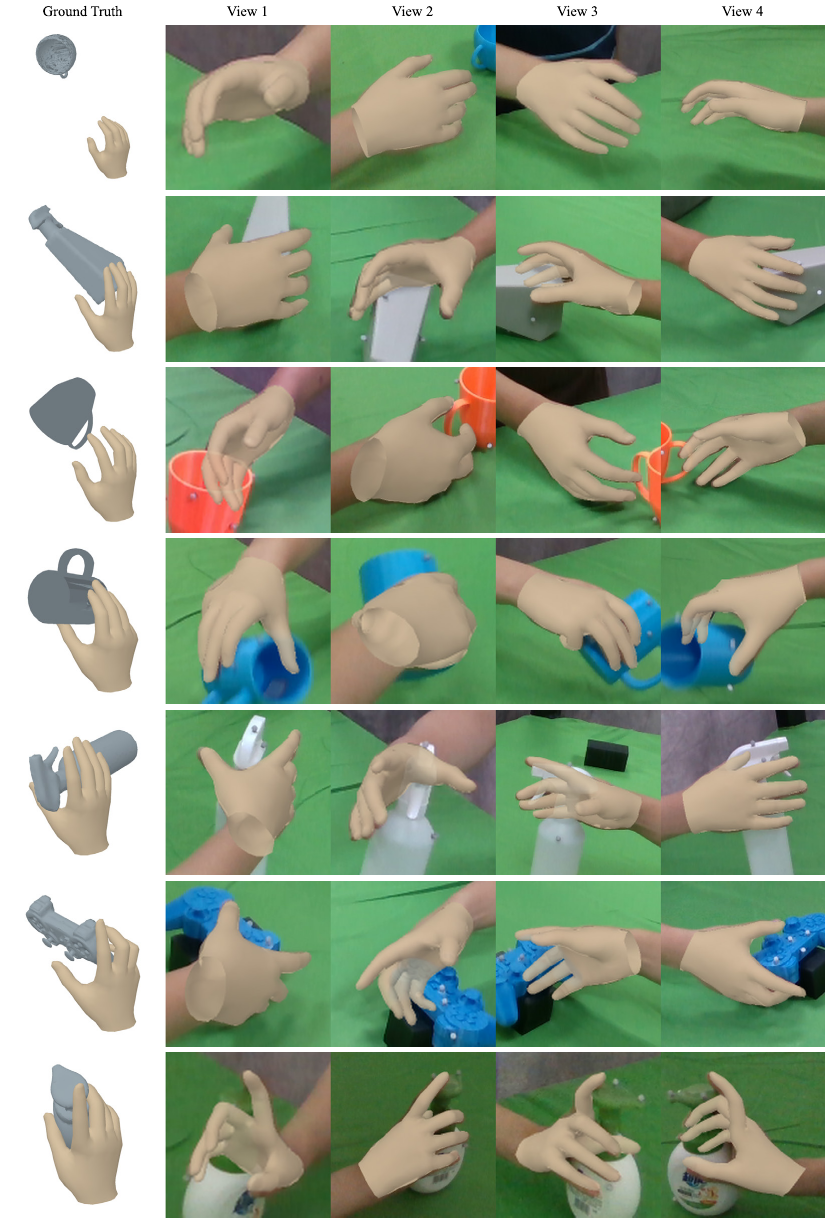}
    \caption{\textbf{Qualitative results on OakInk-MV} with SAM3-predicted masks as input.}
\end{figure}

\begin{figure}[!t]
    \centering
    \includegraphics[width=\linewidth]{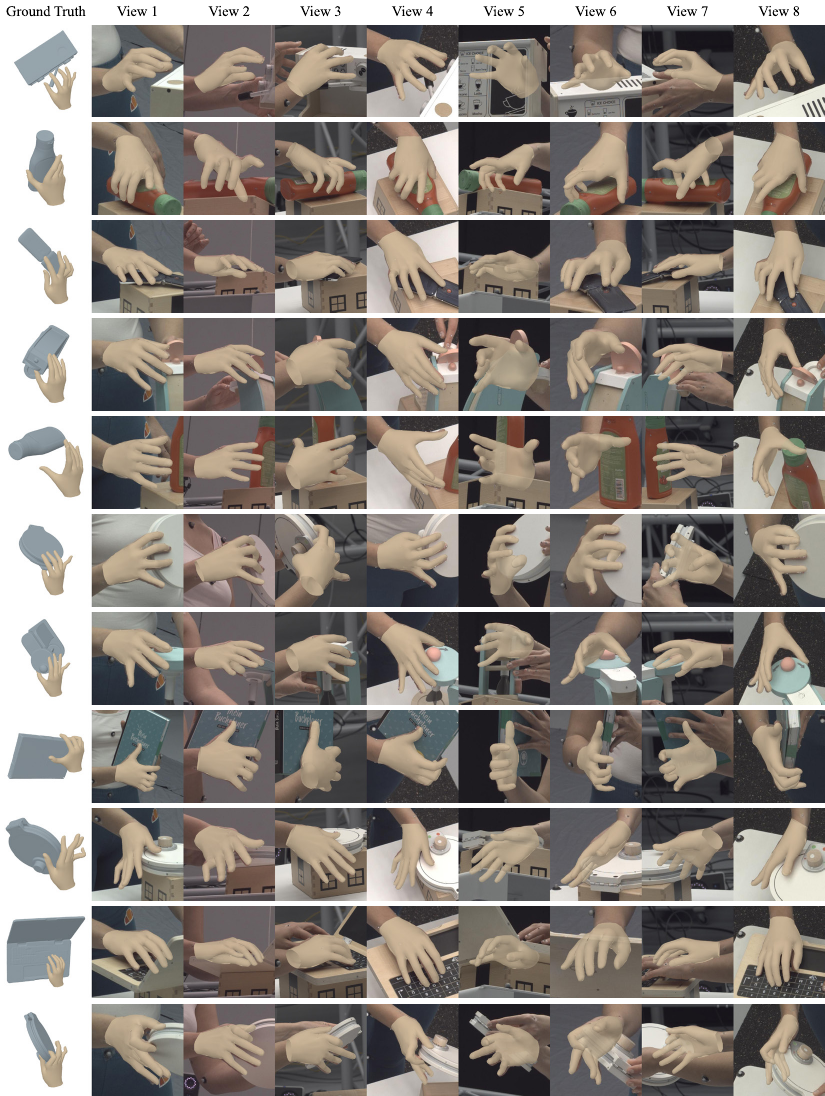}
    \caption{\textbf{Qualitative results on ARCTIC-MV} with SAM3-predicted masks as input.}
\end{figure}

\begin{figure}[!t]
    \centering
    \includegraphics[width=\linewidth]{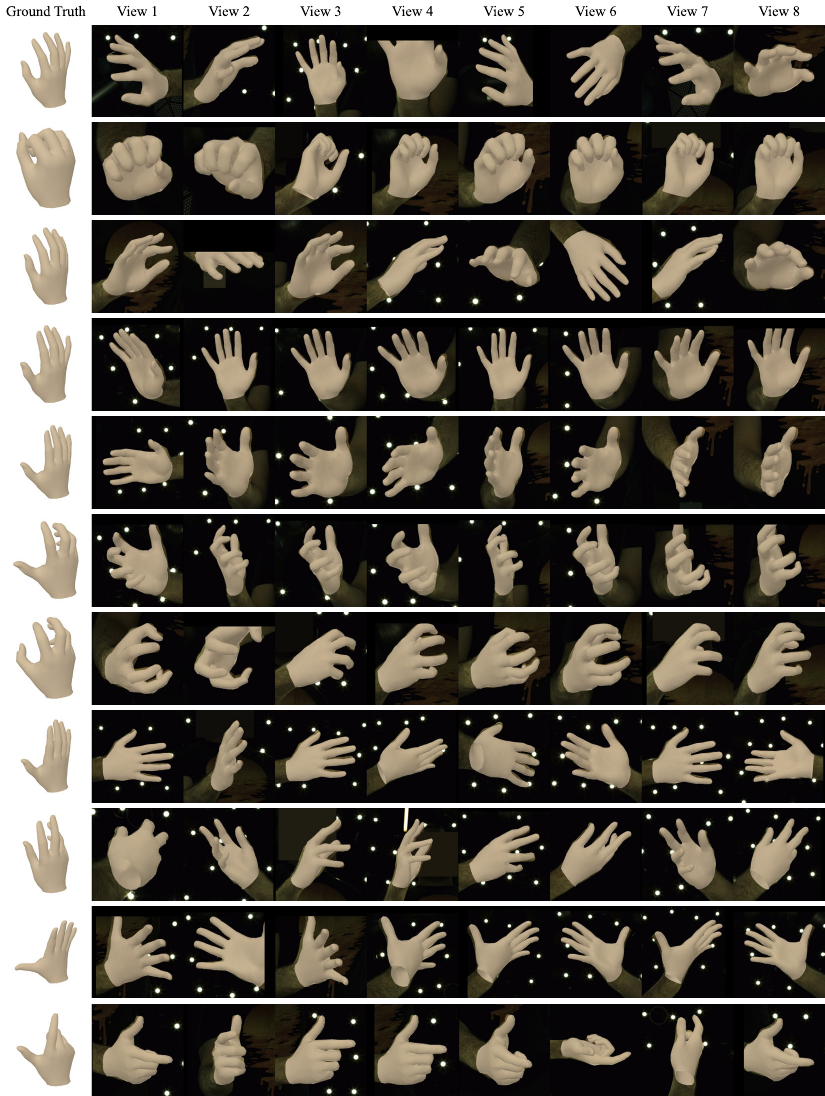}
    \caption{\textbf{Qualitative results on InterHand-MV} with SAM3-predicted masks as input.}
\end{figure}

\end{document}